\documentclass[11pt]{article} 
\usepackage[utf8]{inputenc} 
\usepackage[T1]{fontenc}
\usepackage{lmodern}
\usepackage[letterpaper, margin=1in]{geometry}
\usepackage{graphicx} 
\usepackage{amsmath, amssymb} 
\usepackage{booktabs}        
\usepackage{cite}            
\usepackage{url}             
\usepackage{microtype}       
\usepackage{hyphenat}
\usepackage{multirow}
\usepackage{float}
\usepackage{subcaption}
\usepackage{url}
\usepackage{authblk}
\usepackage[colorlinks=true, allcolors=blue]{hyperref}
\usepackage{cleveref}

\providecommand{\keywords}[1]{\textbf{Keywords:} #1}

\title{The Singularity Space: A Generative Diffusion Framework for Signal Representation}

\author[1]{Eli Bar-Yosef \thanks{Email: \texttt{elibaryosef@mail.tau.ac.il}}}
\affil[1]{Department of Applied Mathematics, Tel Aviv University, Israel}
\author[2]{Amir Averbuch \thanks{Email: \texttt{amir@math.tau.ac.il}}}
\affil[2]{School of Computer Science and AI, Tel Aviv University, Israel}
\author[1]{Eli Turkel \thanks{Email: \texttt{turkel@tauex.tau.ac.il}}}
\date{}
\numberwithin{equation}{section}
\numberwithin{figure}{section}
\numberwithin{table}{section}

\begin{document}
\maketitle

\begin{abstract}
Generative models often represent signals as dense grids of amplitudes, blurring sharp transients that are crucial for the correctness of physical signals. We introduce \emph{Singularity Space}, a generative framework that represents signals through complex-plane singularities, rooted in the classical pole-residue representation of meromorphic functions. We learn a latent space of physically constrained, per-signal singularity configurations to solve an inverse problem from degraded or partial observations. The framework has three key properties: interpretability, in which each generated singularity configuration corresponds to a set of physical parameters; structural stability, which mitigates Gibbs artifacts at discontinuities; and resolution-free output reconstruction on arbitrary grids without retraining or interpolation. Our framework employs a transformer-based diffusion model that directly predicts samples at complex-plane singularity coordinates, subject to geometric constraints during sampling. As a controlled test case for sharp-feature recovery, we evaluate our framework on 1D Burgers shocks, where each shock is represented by 32 predicted singularities (an $8\times$ reduction versus a 1024-point grid signal). Our framework preserves signal structure ($\text{TV ratio} \approx 1$) under unseen test-time observation noise, achieves a $4.2\times$ lower reconstruction error in zero-shot sub-resolution generalization than a grid-based baseline, and recovers physical parameters to $10^{-4}$ absolute error in-distribution. These results suggest that singularity-based representations may provide a practical foundation for other transient-dominated signals such as speech and biomedical signals, with potential extension to higher-dimensional domains.
\end{abstract}

\section{Introduction}
Real-world multiscale signals, whether sharp acoustic transients, physiological spikes (e.g., QRS complexes, neural spikes), or discontinuities in fluid dynamics, are often characterized by a localized transient structure. 
Generative models, such as diffusion models and autoencoders, have achieved significant success in natural image and audio synthesis by treating data as values on fixed grid sampled arrays. However, these models are optimized for perceptual fidelity rather than physical accuracy, where the amplitude, location and sharpness of a discontinuity or transient are crucial for physical correctness. Moreover, the grid formulation creates a representational mismatch: the physical phenomenon is localized and transient, yet the network is forced to represent it on a dense grid. To resolve these challenges, modern generative architectures must coordinate many independent amplitudes to approximate a single transient. A small transient misalignment can push grid-based models toward low-frequency, averaged reconstructions, suppressing the high-frequency components needed to accurately represent the signal. This results in non-physical smoothed or blurred transients. This phenomenon is related to spectral bias~\cite{rahaman2019spectralbiasneuralnetworks}, where neural networks converge on low-frequency approximations early during training and are slow to learn high-frequency parts of the signal. These high-frequency parts are crucial for accurately representing sharp transients.
Spectral representations are often used to mitigate the inefficiency of dense grid representations through truncated spectral modes. 
This is effective for smooth signals, where low-frequency modes dominate, but for sharp localized transients these truncated approximations may smooth the transition or produce Gibbs-type oscillations near the jump. Recent state-of-the-art neural operators attempt to fix these issues by relying on spectral or Fourier-based backbones. However, these methods struggle to resolve localized transients. Hybrid methods~\cite{oommen2025integratingneuraloperatorsdiffusion, xyjy-gx6f, Zhang2024} attempt to mitigate these issues, but do not resolve the underlying representation challenge. The inefficiency in capturing sharp transients is not a failure of training, but a limitation of the representation itself. 
To efficiently resolve high-frequency features, we move beyond the grid and adopt a representation that expresses discontinuities and sharp transients through isolated complex-plane singularities. 

We introduce Singularity Space, a generative framework that decouples the structure, represented by the singularity set $\mathcal{V}=\{(z_k,r_k)\}_{k=1}^{N}$, from the grid-valued signal. In this framework, we operate at a higher level of abstraction by exploiting the principle that, in the presence of singularities (up to an analytic background term), every meromorphic signal is uniquely determined by its singularities. Intuitively, if the signal magnitude is viewed as a flexible surface representing its energy landscape, its singularities act as discrete 'tent poles', localized energy sources that anchor and shape the global topology~\cite{ambardar2003digital}.
These structural anchors are discrete; however, the physical field they generate is inherently continuous. Learning the distribution of these singularity points allows bypassing the representational bottlenecks of standard grid-based networks. This abstraction layer shifts the representational burden; instead of the neural network approximating the high-frequency values of a signal, it generates the underlying discrete singularity coordinates. Our framework adopts a nonlinear analytical approach rooted in classical complex analysis~\cite{stein2003complex, coifman2026holomorphic}, so rather than generating grid amplitudes, the model produces a compact, interpretable per-signal representation consisting of a permutation-invariant, complex-plane singularity set $\mathcal{V}$, where $z_k=x_k+iy_k$ denotes the complex-plane singularity location and $r_k$ its associated residue. The generated singularity configuration is directly associated with the physical features and implicitly optimized via the learned singularity distribution, specifically, localizing events via $x_k$ and controlling transient sharpness via $y_k$. Intuitively, our network acts as a low-frequency architect of high-frequency geometry. Thus, our framework can represent not only sharp transients and discontinuities but may also be extendable to smooth signals. This represents a fundamental shift: while standard spectral operators act on a fixed global frequency grid, our model optimizes the nonlinear coordinates of the adaptive meromorphic expansion. Thus, each generated pole $z_k$ defines an adaptive analytic component of the signal. The sharpness (or scale) of this component is dynamically regulated by the generated pole's proximity to the real axis ($1/|y_k|$), rather than by fixed, uniform spectral modes.

\begin{figure}[H]
\centering
\includegraphics[width=\linewidth]{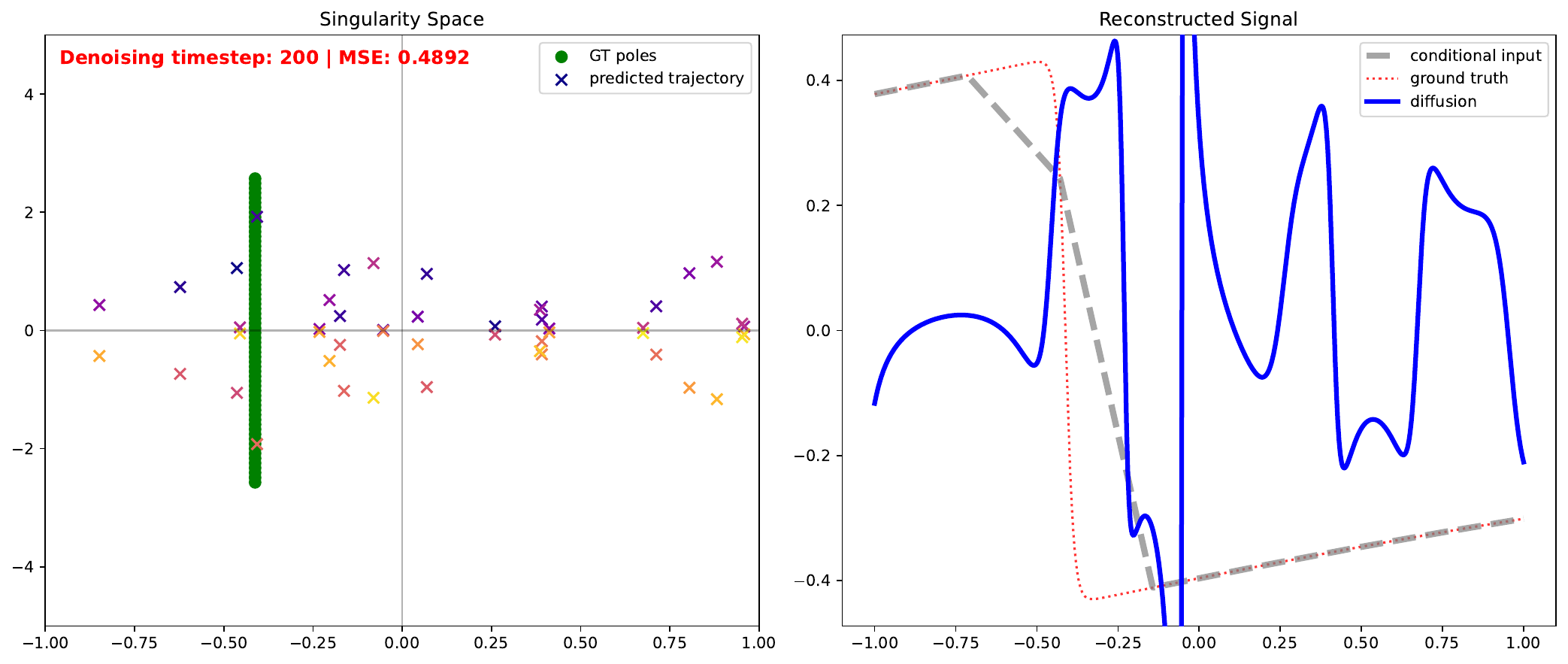}
\includegraphics[width=\linewidth]{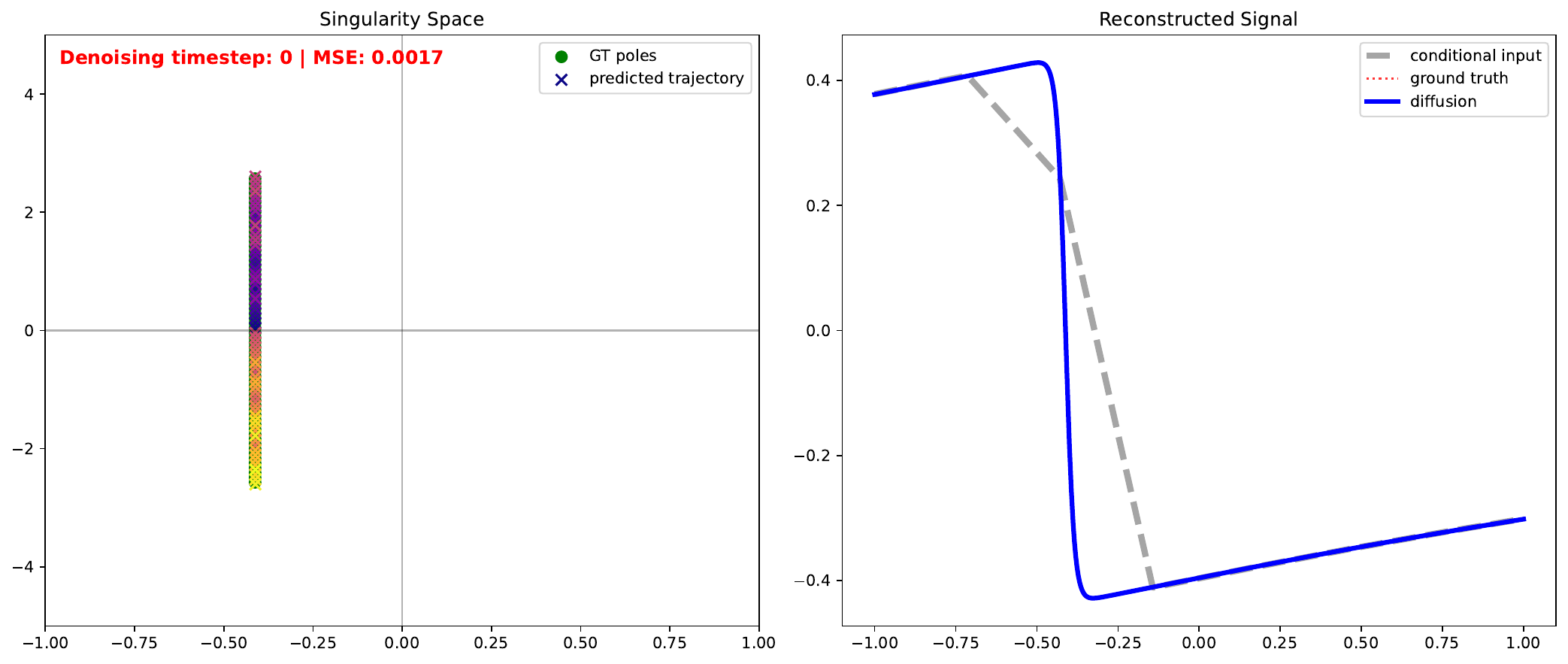}
\caption{Reverse-diffusion steps in Singularity Space, shown at an intermediate step (t=200, top) and at convergence (t=0, bottom). Left: the model denoises a random coordinate state into a valid singularity configuration. Right: the corresponding reconstructed signal (blue) is shown in signal space, conditioned on the observed signal (grey dashed) with the ground-truth signal (red). The scattered singularities converge to the ground-truth configuration, and the reconstruction matches the correct shock.}
\label{fig:diffusion_trajectory}
\end{figure}

To capture the interactions among these singularities, we introduce a Transformer-based diffusion architecture that operates in the complex plane $\mathbb{C}$ and learns the joint distribution over the set of singularities. We utilize a score-based Stochastic Differential Equation (SDE) to steer an initial standard normal distribution, $\mathcal{N}(0, I)$, into physically-consistent singularity configurations. We formulate our denoising objective in terms of direct sample prediction ($\hat{\mathbf{Q}}_0$) rather than the noise ($\boldsymbol{\epsilon}$). This allows us to apply active distance-minimizing projections ($\mathcal{P}_{\Omega_\delta}$) at every discrete step, governing the complex physical constraints during the sampling process. This way, our Singularity Space mitigates spectral bias in the output representation by shifting the learning problem from dense, high-frequency amplitude prediction to low-dimensional singularity generation. This is visually demonstrated in Fig.~\ref{fig:diffusion_trajectory}, where an initial random Gaussian sample is steered by corrupted observation toward a physically valid signal. 

Although spectral bases typically remain optimal for smooth signals, our framework may represent oscillatory smooth dynamics by coordinating singularities farther from the real axis (i.e., with large $|y_n|$), thereby broadening the spatial influence into a smooth, low-frequency signal.
While the conditioning observation is sampled on a grid, the model does not generate a grid-valued output. Instead, the diffusion Transformer generates singularity tokens, which are then reconstructed by a non-learned meromorphic decoder. In Singularity Space, the transient is represented by these generated independent singularities, so a location error appears as an explicit geometric error in the singularity coordinates rather than being distributed across many independent grid amplitudes. 

To evaluate our Singularity Space representation and numerical stability, we choose a domain with an analytical ground truth and well-defined singularity behavior. The non-linear 1D viscous Burgers equation serves as a controlled test case for our singularity representation. We generate synthetic shock data from the analytical solution and formulate the evaluation as an in-distribution, zero-shot inverse problem to recover the original shock from the degraded shock observations in the near-singular limit. Specifically, we demonstrate this by conditioning our diffusion process on degraded and partial shock observations and achieving high-fidelity reconstruction of shocks in low-viscosity settings ($\nu \to 0$).

The potential applicability of our framework extends beyond shocks to other transient-based signals, such as speech and biomedical signals (e.g., ECG/EEG), which all contain localized high-frequency structures embedded in smoother background dynamics. These domains are therefore candidates for future exploration of Singularity Space.

\subsubsection*{Our main contributions are:}
\begin{itemize}
    \item \textbf{The Singularity Space Representation:} We propose a generative framework for signals with a meromorphic representation. Rather than representing signals on discrete grids or fixed spectral bases, we learn the underlying conditional distribution over complex-plane singularities. Each singularity contributes to a localized physical feature (e.g., position, decay rate, and amplitude), so sharp transients are represented through the generated singularity set $\mathcal{V}$ rather than dense high-frequency grid amplitudes. This allows us to generate a compact and interpretable singularity representation that can be reconstructed on arbitrary output grids and mitigates Gibbs oscillations by analytic reconstruction. 
    \item \textbf{Amortized Diffusion over Singularities:} We design a Transformer-based diffusion architecture operating on singularity coordinates in $\mathbb{C}$, using attention to coordinate singularities and condition them on degraded observation. By processing signals as a set of singularities by a direct sample prediction ($\hat{\mathbf{Q}}_0$), we enforce physical correctness through distance-minimizing projections ($\mathcal{P}_{\Omega_\delta}$). This replaces per-instance rational fitting with amortized inference, and enables reconstruction of an ill-posed inverse problem from degraded and partial observations.
    \item \textbf{Resolution-Free and Zero-Shot Generalization:} We evaluate our framework on the 1D viscous Burgers equation in the low-viscosity limit ($ \nu \sim 10^{-3}$). The analytic decoder reconstructs inferred singularity sets at arbitrary output grids without retraining or interpolation. In zero-shot sub-resolution in an underdetermined setting with $M=8$ (half the minimum training resolution), our generative framework achieves a $4.2\times$ lower $L_2$ error, $L_2=0.085$, versus FNO's $L_2=0.360$. The TV-ratio deviation from the ideal value is about $57\times$ smaller than that of FNO. We also show resolution-free, high-precision reconstruction metrics on unseen grids, $G \in \{512, 2048, 4096\}$. 
    \item \textbf{Structural Robustness and Physical Interpretability:} Under test-time noise (without noise-augmented training), our framework maintains lower TV ratios close to the ideal value of $1.0$. FNO's TV ratio grows to $2.654$ at $10\%$ noise, and classical AAA diverges (TV ratio exceeding $100$) at the same noise level. Moreover, the inferred singularity sets are physically interpretable: shock position, viscosity-dependent width, and amplitude are recovered to $10^{-4}$ precision, enabling the direct physical analysis and diagnostics of the generated representation.
\end{itemize}

\section{Related Work}
\label{sec:related_work}

\subsubsection*{Classical Signal Theory and Rational Approximation}
The concept of modeling signals by their singularities has a long history in classical signal processing, where singularity analysis encodes a system's structure in the complex plane and provides a mathematical foundation for filter design and signal processing~\cite{ambardar2003digital, oppenheim1996signals}. 
Classical rational approximation methods, such as AAA~\cite{Nakatsukasa_2018} and Vector Fitting~\cite{vectorfitting_1999}, are analytical per-instance algorithms for approximating functions by extracting pole-residue structure. Rational fitting is sensitive to noise and does not directly encode prior knowledge about physically valid singularity configurations. In contrast, Singularity Space is not about the extraction of poles and residues alone, but the construction of a generative latent space over singularity sets to solve inverse problems.

\subsubsection*{Blaschke Products and Orthogonal Rational Bases}
Signal representation via rational functions has been explored for decades with practical applications. One line of research uses Blaschke products and Malmquist-Takenaka (MT) systems that construct adaptive orthonormal bases for Hardy spaces~\cite{coifman2021multiscaledecompositionshardyspaces,coifman2025practicaluseblaschkedecomposition}. 
The MT basis construction via Blaschke factors introduces a structural coupling in which each basis element depends on all the previous root locations. 
Recent work adapting the Blaschke product to neural networks, such as Blaschke Decomposition Networks~\cite{zhang2025bdn}, utilizes the unwinding series 
$F_L = \sum_{k=0}^{L} c_k \prod_{j=1}^{k} B_j$. This sequential hierarchy formulation demonstrates the structural constraints on every term (excluding the first) of the expansion, where $\prod_{j=1}^k B_j$ is a function of all preceding root locations $\{z_1, \dots, z_{k}\}$. 
When using these mathematical objects in deep learning, typically a dense Jacobian is produced during optimization. Specifically, as the authors noted, this introduces various numerical instabilities in neural networks (e.g., multiplicative overflow, gradient vanishing). Even though the authors address these numerical engineering instabilities in their work, the underlying sequential coupling still persists. Our framework avoids a sequential Blaschke unwinding product. Instead, it adopts an additive meromorphic expansion parametrized on the singularities. This enables a parallel optimization process that relies on a Transformer-based diffusion model, in which each singularity is an independent token that can be generated and optimized in parallel.

\subsubsection*{Spectral Operators and Pole-Residue Parameterizations}
Neural operators are a class of neural networks that learn mappings between function spaces, by learning an operator that maps one grid value function to another. 
Architectures such as Fourier Neural Operator (FNO) represent functions by internally operating on a fixed, truncated set of Fourier modes~\cite{li2021fourierneuraloperatorparametric}. FNO has been successful for smooth parametric Partial Differential Equation (PDE) dynamics, but a spectral truncation is inefficient for localized transients where the signal's spectral energy extends beyond the truncated wavenumber $K$. The frequency cutoff $K$ acts as a low-pass filter and is recommended to be small (e.g., FNO's default $K \le 16$) to maintain computational tractability~\cite{li2021fourierneuraloperatorparametric}. Although Li et al. suggest that a local non-linear activation can recover these discarded high-frequency modes \cite{li2021fourierneuraloperatorparametric}, the activation alone may be insufficient to keep the accuracy of localized transients. Sharp transients require many coordinated high-frequency modes for accurate localization, making them less natural for such representations. Hybrid methods~\cite{oommen2025integratingneuraloperatorsdiffusion, xyjy-gx6f, Zhang2024} attempt to reintroduce these high-frequency components, for example, by integrating the diffusion-based generated components. However, they remain limited by the underlying spectral design. We address these limitations in the Singularity Space by directly parameterizing the underlying singularities. Instead of fixed frequency-domain truncation, we shift to generative, adaptive coordinates in the complex plane. 

Laplace Neural Operator (LNO)~\cite{cao2023lnolaplaceneuraloperator} takes a related but distinct approach. It uses a poles-residue parametrization of the neural operator kernel. This demonstrates that a complex-plane structure can improve transient modeling. 
However, LNO remains a deterministic operator-learning framework, in which its poles are neural operator-level parameters shared across inputs, its per-signal output is a dense field, and its interpretability is at the operator/system level rather than per reconstructed signal. 

In contrast, the Singularity Space representation treats singularities as the inferred output representation. Thus, given a degraded observation, the model generates a per-signal singularity configuration under a learned prior and decodes it analytically. Practically, we shift the representation from a fixed frequency-domain truncation or operator-level pole parameterization to adaptive, generative coordinates in the complex plane.

\subsubsection*{Coordinate-Based Representations and Physics-Informed Networks}
Coordinate-based networks represent signals as functions of continuous coordinates by learning a map from query locations to signal values. 
These networks have been adopted across domains, from visual scene representation to physical system modeling. SIRENs~\cite{sitzmann2020implicitneuralrepresentationsperiodic} for a continuous signal representation (i.e., audio, video, and images), NeRFs~\cite{mildenhall2020nerfrepresentingscenesneural} for vision, and Physical Informed Neural Networks (PINN)~\cite{PINNRAISSI2019686} for physical systems share a common approach: they map spatial or temporal coordinates to continuous signal values. They are capable of resolution-free reconstruction; however, they often rely on computationally expensive per-instance fitting or problem-specific residual losses (e.g., PINN). Furthermore, in these architectures, coordinates are "passive" inputs to an MLP that directly maps $x$ to a signal value $u(x)$, without a relation between parameters and physical quantities. 

For example, PINNs approximate the solution by minimizing a composite loss function, consisting of specific solution objectives, such as physical residual, initial, and boundary conditions~\cite{hyde2026PINNs}. These objective functions provide a flexible mechanism for integrating physical constraints but may limit generalization across instances and require retraining the model~\cite{hyde2026PINNs}. As such, resolving sharp transients remains a well-documented challenge. The globally smooth activation functions and their soft constraint loss cause these networks to smear localized shocks rather than capturing steep gradients~\cite{krishnapriyan2021characterizingpossiblefailuremodes}. 

Singularity Space is continuous and outputs a grid-free representation, but the role of coordinates differs from that in standard coordinate-based networks. Instead of learning a neural map directly, singularity coordinates serve as evaluation points for an additive meromorphic model whose parameters are the singularity locations and residues. These parameters directly encode the physical structure of the signal (e.g., location, scale). By learning a distribution over this interpretable parameter space with a Transformer-based diffusion model, our approach separates representation learning from per-instance coordinate fitting or problem-specific optimization. Sampling a new reconstruction requires only a fixed number of reverse denoising steps, after which the generated pole-residue set can be evaluated analytically at arbitrary coordinates. 

\section{Proposed Methodology}

\subsection{The Singularity Space Representation}
Physical signals are typically generated by specific dynamics. Their information is often concentrated at localized transients such as shock fronts or ECG QRS complexes, which are inefficiently captured by grids or global bases~\cite{rahaman2019spectralbiasneuralnetworks}.
We propose that signals with sharp transients can be efficiently represented in the complex plane and decomposed into an interpretable, compact, and adaptive rational representation. In this representation, a set of isolated singular points, $z_k\in\mathbb{C}$, with associated residues $r_k$ at the corresponding complex coordinates encodes the dominant analytic structure of the signal. For instance, sharp acoustic shocks or neuronal action potentials are modeled using a dense pixel grid or a global sum of sine waves. However, in the complex plane, they can be represented by a small discrete set of mathematical singularities. We take this idea further and define a generative space based on these principles, which we refer to as the Singularity Space. To mathematically formalize our generative Singularity Space, we restrict our generative space to the meromorphic function space $\mathcal{M}(\mathbb{C})$. A meromorphic function is a complex-valued function that is holomorphic everywhere, that is, differentiable in an $\epsilon$-neighborhood of each point in its domain, except at isolated poles~\cite{stein2003complex}. We define a singularity as a point at which a function fails to be holomorphic. In this work, we only consider points of pole singularities; specifically, we refer to a singularity as an isolated pole unless stated otherwise. This excludes essential singularities, where the function assumes every possible value in a small neighborhood, or branch point singularities, whose multi-valued behavior creates representational ambiguity. Furthermore, while the meromorphic space $\mathcal{M}(\mathbb{C})$ may contain functions with an infinite set of isolated singularities, physical observations are limited by finite resolution and bandwidth. By restricting this meromorphic class to a finite-pole rational approximation, we get a compact and tractable representation given an observed signal.

\subsection{Problem Formulation} We define a real signal $u(x)$ over a finite domain as a meromorphic function $f(z) \in \mathcal{M}(\mathbb{C})$, where $f(z)$ is defined as the ratio of two polynomials: 
\begin{equation}
\label{eq:rational_ratio}
f(z) = \frac{P(z)}{Q(z)}. 
\end{equation}
We require the system to be strictly proper by enforcing $\deg(Q) > \deg(P)$, where $\deg(Q) = N$ and assume the roots of the denominator are simple (i.e., multiplicity of one). These constraints ensure that our representation avoids polynomial remainders and higher-order poles, thereby keeping the system physically correct and unambiguous for learning.

Following the Fundamental Theorem of Algebra~\cite{churchill2013complex}, the denominator $Q(z)$ is uniquely determined by its $N$ complex roots, which determine the pole locations of $f(z)$. As such, $f(z)$ can be parametrized by a finite set of $N$ isolated singularities, denoted as:
\begin{equation}
\label{eq:singular_set}
\mathcal{V} = \{(z_k, r_k)\}_{k=1}^N, 
\end{equation}
where $z_k \in \mathbb{C}$ is the $k$-th pole with its corresponding $k$-th residue $r_k \in \mathbb{C}$.
At each singularity pair $(z_k, r_k)$, where $z_k, r_k \in \mathbb{C}$, $z_k$ denotes the frequency and decay rate, and the residue $r_k$ defines the amplitude and initial phase of each physical mode. To ensure that the singularity set $\mathcal{V}$ represents a real signal $u(x) \in \mathbb{R}$, we enforce conjugate symmetry, so that for each singularity pair $(z_k, r_k)$ where $\text{Im}(z_k) \neq 0$, we include the conjugate symmetry set $(z_k^*, r_k^*)$ as part of the set $\mathcal{V}$.

\subsection{Additive Singularity Representation}
\label{sec:additive_rep}
The rational form $f(z)$ in Eq.~\ref{eq:rational_ratio} represents a real signal $u(x)$ as a meromorphic function; this ratio of two polynomials provides a global, highly coupled parameterization of the system. In this parametrization, small perturbations to the polynomial coefficients can simultaneously move multiple roots.
To ensure stable generative modeling, we require an additive representation in which singular components are parameterized directly. 
To obtain such a representation, we follow the finite-pole form of the Mittag-Leffler expansion~\cite{stein2003complex}. The meromorphic function can be written as a sum of principal parts at its poles plus an entire function:
\begin{equation}
\label{eq:add_rep_mittag_full}
f(z) = \sum_{k=1}^{N} p(z; z_k, r_k) + h(z),
\end{equation}
where $p(z; z_k, r_k) = \frac{r_k}{z - z_k}$ is the principal part of the Laurent series expansion centered at the pole $z_k$, where each $z_k$ and $r_k$ are included in the singularity set $\mathcal{V}$ of size $N$ (Eq.~\ref{eq:singular_set}). 
The arbitrary holomorphic function, $h(z)$, represents global trends and is assumed to be zero due to the enforced strictly proper constraints. Future work could extend this framework, superimposing singularities onto smooth or oscillatory backgrounds (e.g., a sharp shock over a sine wave) by parameterizing $h(z)$ as an additional analytic background component. This separates the singular structure from the globally smooth component. In this work, where $h(z) = 0$, the relationship between the real signal $u(x)$ in $\mathbb{R}$ and the singularity set $\mathcal{V} \subset \mathbb{C}$ is defined as follows:
\begin{equation}
\label{eq:add_rep_mittag_part}
u(x) = \sum_{(z, r) \in \mathcal{V}} \frac{r}{x - z}.
\end{equation}
The additive structure in Eq.~\ref{eq:add_rep_mittag_part} defines an adaptive rational expansion over the conjugate symmetric singularity set $\mathcal{V}$, where the function of each singular component, $\phi_k(x) = (x-z_k)^{-1}$, is not fixed but is dynamically parameterized by the data. Eq.~\ref{eq:add_rep_mittag_part} also defines the fixed meromorphic decoder used throughout this work. We denote this decoder by $\mathcal{D}$. Thus, given a generated singularity set $\mathcal{V}$, it evaluates the singularity set on a chosen real grid to produce the reconstructed signal. Moreover, changing one pole affects only its own singular component in Singularity Space. This ensures that the influence of each singularity $z_k$ and its corresponding residue $r_k$ retains its effective locality such that the model can adjust a single singularity without destroying the global structure. This also results in a decoupled gradient during optimization in Singularity Space, where cross-derivatives between independent singularities are zero (i.e., $\frac{\partial^2 u}{\partial z_i \partial z_j} = 0$ for $i \neq j$) and avoids gradient coupling as in the case of a sequential structure of product-based expansions, where basis functions depend on products of earlier factors. Furthermore, this parameterization avoids the numerical sensitivity associated with recovering roots from polynomial coefficients, as seen in Wilkinson-type phenomena, where small perturbations in the coefficients can produce large changes in the corresponding roots. Rather than generating polynomial coefficients and then extracting roots, our model generates pole locations and residues directly in the complex plane. The direct coordinate parametrization is important for our generative process because the diffusion model can learn independent trajectories for distinct singularities within the complex plane. 
Eq.~\ref{eq:add_rep_mittag_part} also suggests why the representation can remain compact. The required number of singular components is expected to scale with the number and complexity of localized analytic structures, rather than directly with the sampling grid resolution or maximum spectral frequency. This decouples the representation from the grid and allows sparse, high-frequency events to be represented using a compact set of adaptive complex-plane coordinates, without the parameter growth typical of spectral or grid-based representations. 

We view the singularity framework as a general representation for transient-dominated signals. Mathematically, this representation is inherently designed to support meromorphic functions with finite isolated poles. Since we are concerned with finite physical domains, this representation can also be extended to smooth components by coordinating singularities in the complex plane or, as previously noted, by utilizing the holomorphic term $h(z)$ (Eq.~\ref{eq:add_rep_mittag_full}).

\subsection{Coordinate Space for Singularity Sets} 
A signal represented compactly through a singularity set $\mathcal{V}$ (Eq.~\ref{eq:singular_set}) forms a finite-dimensional latent parameter space, its $N$ upper-half-plane singularities lie in a subset of $\mathbb{C}^{2N}\cong\mathbb{R}^{4N}$. Instead of learning a dense collection of amplitudes (e.g., pixels), as in grid-based approaches, we use a singularity-coordinate parameterization, in which each singularity-coordinate encodes interpretable physical information. Generating valid singularity sets requires a stable coordinate system. As shown in Sec.~\ref{sec:additive_rep}, our formulation decouples the singularity set into additive principal modes for numerical stability. Generating a physical singularity set that represents a signal requires high-precision generation of these parameters. A small perturbation in our singularity set, specifically, as $y_k \to 0$, results in $1/|x - z_k|$ diverging locally, resulting in a non-linear, hyperbolic change in the reconstructed signal. For example, the derivative of $\frac{1}{y}$ is $-\frac{1}{y^2}$. At $y=0.1$, the sensitivity is $10^2$. At $y=0.001$, the sensitivity is $10^6$. To mitigate this hyperbolic sensitivity as $y \to 0$, we map the $N$ upper-half-plane singularities of $\mathcal{V}$ into unconstrained log-space. We define the bijective coordinate map $\phi:\mathbb{H}\times\mathbb{C}\to\mathbb{R}^4$, where $\mathbb{H} = \{z \in \mathbb{C} \mid \operatorname{Im}(z) > 0\}$, mapping $(z_k,r_k)$ pair to log-space coordinate:
\begin{equation}
\label{eq:phi_log_space}
\mathbf{q}_k \triangleq \phi(z_k,r_k)=\left[\operatorname{Re}(z_k),\log(\text{Im}(z_k)),\operatorname{Re}(r_k),\,\operatorname{Im}(r_k)\right]\in\mathbb{R}^4, 
\end{equation}
where the corresponding inverse map recovers the physical singularity set as $z_k=q_{k,1}+i\exp(q_{k,2})$, which guarantees $\operatorname{Im}(z_k)>0$, while the residue is recovered as $r_k=q_{k,3}+i q_{k,4}$. 
Collecting these coordinates, we denote the full representation as $$\mathcal{Q} = \{\mathbf{q}_k\}_{k=1}^N, \quad \mathbf{q}_k \in \mathbb{R}^4.$$
The primary purpose of this map (Eq.~\ref{eq:phi_log_space}) is to enforce the upper-half-plane constraint by taking $y_k \to 0$ to $\log(y_k) \to -\infty$ and to ensure that perturbations in the latent singularities $\mathcal{Q}$ result in improved sensitivity in the physical signal $u(x)$. 
This formulation allows the stochastic diffusion process to operate over the unconstrained real-valued singularity coordinates $\mathcal{Q}$. 
Thus, we define a general analytic parameterization for signals that can be represented by isolated singularities. 

\subsubsection{Diffusion over Singularity Coordinates}
\label{sec:diffusion}
We learn the distribution over the normalized log-space singularity coordinates, $\mathcal{Q} \in \mathbb{R}^{4N}$, using a diffusion process~\cite{song2021scorebasedgenerativemodelingstochastic}.
This generative process, in contrast to standard generative architectures (e.g., feed-forward regression or one-shot VAEs), does not predict the entire $4N$-dimensional state $\mathcal{Q}$ in a single forward pass. By spreading coordinate generation across $T$ iterative refinement steps, the model avoids resolving the complex topology immediately and instead gradually steers an initial random distribution toward a physically correct configuration.

We start by describing the underlying forward process using a continuous-time SDE formulation that gradually perturbs the initial singularity set $\mathcal{Q}_0 \sim p_{\mathrm{data}}$ into $\mathcal{Q}_T \sim \mathcal{N}(0, I)$. We define the standard Variance Preserving forward SDE as 
\begin{equation}
\label{eq:diffusion_forward}
d\mathcal{Q}_t=f(\mathcal{Q}_t, t)dt + g(t)d\mathbf{w}_t, \qquad t \in [0, T], 
\end{equation}
where $f(\mathcal{Q}_t, t) = -\frac{1}{2}\beta(t)\mathcal{Q}_t$ is the drift coefficient, $g(t) = \sqrt{\beta(t)}$ is the diffusion coefficient controlled by the noise schedule $\beta(t)$, and $\mathbf{w}_t$ is a standard Wiener process. Our objective is to learn the conditional score function $s_\theta(\mathcal{Q}_t, t, \mathbf{c}) \approx \nabla_{\mathcal{Q}_t} \log p_{t}(\mathcal{Q}_t \mid \mathbf{c})$, where $\mathbf{c}$ is an observational vector, such as a degraded or corrupted observation. Conditioning the generative process on observation $\mathbf{c}$ enables the model to learn a conditional distribution $p(\mathcal{Q}_0\mid \mathbf{c})$ over valid latent singularity configurations that are consistent with the observed signal.

Standard diffusion models typically operate in unconstrained domains; however, in our singularity domains, if two singularities collide/overlap, our mathematical assumptions are violated. This makes the singular representation unstable.
Let $\zeta(q_i)=q_{i,1}+i\exp(q_{i,2})$, which maps the $q_i$ to its pole location. 
The singular representation becomes invalid when $|\zeta(q_i)-\zeta(q_j)|\to 0, \qquad i\neq j.$
Specifically, the two singularities merge into a single higher-order pole, violating the simple-pole assumption. 
To prevent these collisions, we define the reverse generative process as a constrained conditional generation under the Projected Diffusion Models (PDM) framework~\cite{christopher2024constrainedsynthesisprojecteddiffusion}:
$$d\mathcal{Q}_t = \left[\mathbf{f}(\mathcal{Q}_t, t) - {g(t)}^2 s_\theta(\mathcal{Q}_t, t, \mathbf{c}) \right] dt + g(t)d\bar{\mathbf{w}}_t,$$
where $d\bar{w}_t$ is a backward-time Wiener process. The conditional observation $\mathbf{c}$ guides the generative process to produce a singularity configuration that recovers (or completes) the observed data. The constraint $\mathcal{Q}_t \in \Omega_{\delta}$ is iteratively enforced during the reverse generative process via a projection operator defined as a distance minimization problem, 

\begin{equation}
\label{eq:enforce_op}
\begin{aligned}
\mathcal{P}_{\Omega_{\delta}}(\tilde{\mathcal{Q}}_t) &= \operatorname*{arg\,min}_{\mathcal{Q}'\in\Omega_{\delta}} \sum_{i=1}^{N} \|\mathbf{q}'_{i} - \tilde{\mathbf{q}}_{{t}, {i}}\|_2^2 , \\
\text{subject to:} \quad \Omega_{\delta} &= \{\mathcal{Q}' : \|\zeta(\mathbf{q}'_{i}) - \zeta(\mathbf{q}'_{j})\| \ge \delta, \quad \forall i \neq j\}, 
\end{aligned}
\end{equation}
where $\tilde{\mathcal{Q}}_t$ is the unconstrained singularity set at time $t$, and $\mathcal{Q'}_{t}$ is the optimal candidate singularity configuration at time $t$. This formulation ensures that the resulting configuration $\mathcal{Q}'_t$ maintains a pairwise separation of at least $\delta$ between all the singularities, while remaining the closest possible valid state to the unconstrained $\tilde{\mathcal{Q}}_t$.
Practically, this reverse process is implemented via a discrete Denoising Diffusion Probabilistic Model (DDPM)~\cite{ho2020denoisingddpm} approximation over $T$ refinement steps. 
Unlike standard DDPM formulations, which parameterize the neural network to predict the injected noise, $\boldsymbol{\epsilon}$, our framework is parameterized to predict the clean, network-predicted coordinate state, $\hat{\mathcal{Q}}_0$. These clean coordinate predictions are equivalent to reparameterizing the score-based estimate through the DDPM formula:
$$s_\theta(\mathcal{Q}_t, t, \mathbf{c}) = \frac{\sqrt{\bar{\alpha}_t}\hat{\mathcal{Q}}_0(\mathcal{Q}_t, t, \mathbf{c}) - \mathcal{Q}_t}{1 - \bar{\alpha}_t}.$$
This way, our network predicts the singular coordinates at every timestep, allowing us to verify the singularities along the trajectories and apply the projection $\mathcal{P}_{\Omega_{\delta}}$ to enforce validity. Hard constraints, such as those in standard PDMs, introduce discontinuities in the score function, making deterministic ODE solvers (e.g., DDIM) unstable or trapped on non-convex boundaries~\cite{christopher2024constrainedsynthesisprojecteddiffusion}. The stochasticity of the DDPM reverse step may help the exploration to escape local minima while reaching a stable, non-colliding singularity configuration. Other methods, such as Reflected Diffusion Models~\cite{lou2023reflected}, incorporate boundary constraints into the forward SDE (Eq.~\ref{eq:diffusion_forward}) during training to avoid such artifacts. However, as noted by Lou et al. ~\cite{lou2023reflected}, directly computing transition probabilities on a non-convex manifold with dependent boundaries (e.g., shifting one boundary affects the positions of the others) is computationally intractable. Solving this would let us use deterministic ODE-based samplers (e.g., DDIM), potentially as an extension for future work.

\subsection{Singularity Transformer-Based Denoising Network}
The score function, $s_\theta(\mathcal{Q}_t, t, \mathbf{c}) \approx \nabla_{\mathcal{Q}_t} \log p_t(\mathcal{Q}_t \mid \mathbf{c})$, is a $4N$-dimensional vector field over a continuous coordinate manifold of $N$ interacting isolated singularities. Standard DDPM architectures typically rely on grid-based (e.g., uniformly sampled spatial domains) networks, such as U-Net-based architectures, which are less suited to the Singularity Space topology.
We parameterize the reverse diffusion model using a Singularity Diffusion Transformer (DiT) for predicting a clean state, $\hat{\mathcal{Q}}_0$. To match the singularity grid-less manifold topology, we designed a Transformer-based diffusion model for unordered coordinate sets, where each singularity is represented by an independent transformer token.
The $N$ singularities represent features in the complex plane; however, to interface these complex-valued features with the standard frameworks, each pole $z_k \in \mathbb{C}$ and its associated residue $r_k \in \mathbb{C}$ are represented by the real-valued feature vector $\mathbf q_k \in \mathbb{R}^4$ previously defined in Eq.~\ref{eq:phi_log_space}. 
This representation ensures compatibility with hardware-optimized neural network kernels and avoids the engineering complexity of complex-valued backpropagation. 

\begin{figure}[H]
    \centering
    \includegraphics[page=1,width=0.80\linewidth]{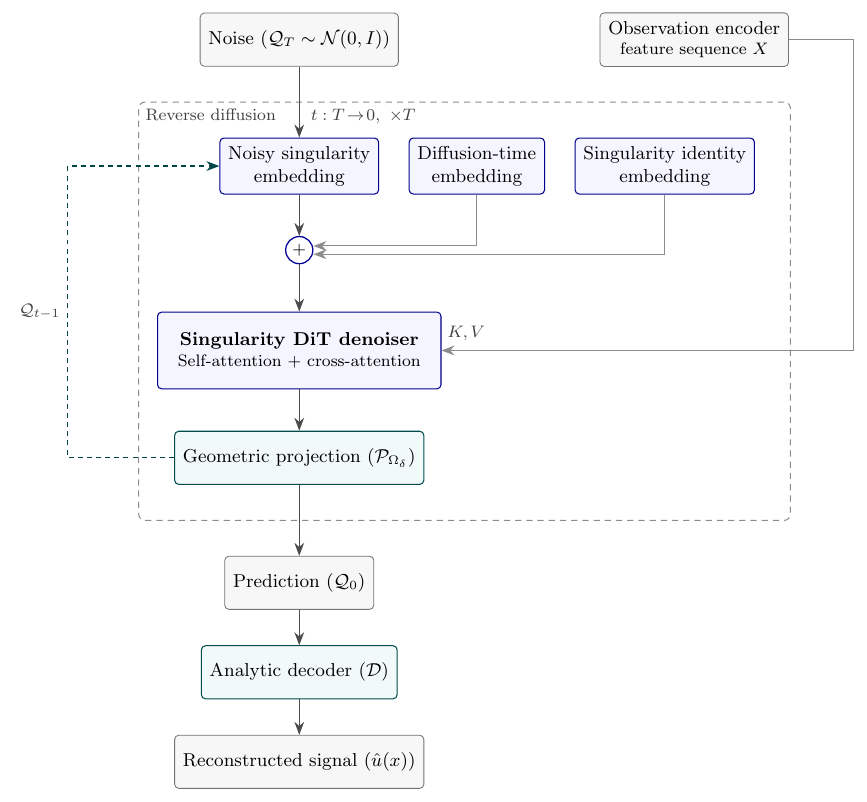}
    \caption{The inference pipeline of the Singularity Space architecture. A degraded observation steers a reverse diffusion process from pure noise into a valid singularity configuration. The denoised singularities are then decoded to reconstruct the signal.}
    \label{fig:singularity_dit}
\end{figure}

The inference pipeline is shown in Fig.~\ref{fig:singularity_dit}: starting from pure noise in the singularity coordinate space, the reverse diffusion process iteratively denoises the coordinates, conditioned on the encoded observation via cross-attention and constrained by a geometric projection, until a clean singularity configuration set $\mathcal{Q}_0$ is predicted and decoded analytically into the reconstructed signal $\hat{u}(x)$.
The Singularity Space DiT design is based on several core choices:

\subsubsection*{Identity Tracking and Permutation Invariance}
The representation of a physical signal through its $N$ singularities is formally defined as a permutation-invariant, unordered set 
$$\mathcal{Q} = \{\mathbf{q}_k\}_{k=1}^N, \quad \mathbf{q}_k \in \mathbb{R}^4.$$ 
Self-attention is architecturally permutation-equivariant, with no built-in token identity. We empirically found that when noisy singularities are statistically indistinguishable, all tokens converge to an averaged coordinate during early diffusion timesteps. To ensure stable tracking on these $N$ unique singularity interactions, the transformer needs a mechanism to differentiate individual tokens during diffusion. 

Let $\mathbf{H}^{(0)} \in \mathbb{R}^{N \times d_{\text{model}}}$ be the initial latent state of the network, where each row tracks a distinct singularity trajectory. 
We define $N$ learnable identity embeddings, $\mathbf{E} = \{\mathbf{e}_k\}_{k=1}^N$, where $\mathbf{e}_k \in \mathbb{R}^{d_{\text{model}}}$. These embeddings are added to the projected singularity vectors, denoted as $\mathbf{h}_{k}^{(0)}$ in $\mathbf{H}^{(0)}$, at the input of each transformer token. Each token is initialized with a unique identity embedding, allowing the model to track $N$ distinct singularity trajectories simultaneously. More specifically, at each timestep $t$, the noisy state of the $k$-th singularity is projected and summed with its learnable identity embedding, $\mathbf{e}_k$, and a global temporal embedding $\mathbf{t}_{emb}$ as follows:
$$\mathbf{h}_{k}^{(0)} = \operatorname{Proj}(\mathbf{q}_{k,t}) + \mathbf{e}_k + \mathbf{t}_{\text{emb}},$$ 
where $\operatorname{Proj}(\cdot)$ is a linear projection, mapping the singularities from $\mathbb{R}^4$ to the latent dimension of size $d_{\text{model}}$, and $\mathbf{t}_{\text{emb}} \in \mathbb{R}^{d_{\text{model}}}$ is derived from a standard sinusoidal embedding of the timestep $t$. Using these latent queries, $\mathbf{h}_{k}^{(0)}$, the network can distinguish individual singularities across timesteps without restricting itself to a specific singularity order in the underlying physical signal.
Despite these learnable identity embeddings, the permutation invariance of the physical representation is preserved through the unordered-set formulation and the matching loss.
Furthermore, we apply a pre-layer normalization to the latent singularity features before self-attention blocks. The additive time and identity embeddings are injected post-normalization. This sequence preserves the gradients that would otherwise be partially absorbed (or vanish) by the pre-normalization layer, thereby weakening the network's ability to track these signals.

\subsubsection*{Singularity Coordination via Self Attention}
We model $N$-spatial singularity interaction via the Multi-Head Self Attention (MHSA) mechanism~\cite{vaswani2017attention}. Through this mechanism, our network learns latent dependencies among singularities and captures physically meaningful interaction patterns in singularity space. The multiple heads let it model several such patterns simultaneously, each in its own subspace. Let $\mathbf{H}^{(0)}$ be the initial matrix of embedded singularities built by stacking the individual singularity states $\mathbf{h}_{k}^{(0)}$. Let $\mathbf{H}^{(l-1)} \in \mathbb{R}^{N \times d_{\text{model}}}$ be the latent matrix as the input to layer $l$, where each row $k$ represents the unique composite latent state of the $k$-th singularity. For a given self-attention head, $m \in \{1, \dots, M\}$, we project the input matrix $\mathbf{H}^{(l-1)}$ into the query ($\mathbf{Q}_m$), key ($\mathbf{K}_m$), and value ($\mathbf{V}_m$) subspaces:
\begin{equation}
\label{eq:sa_qkv}
\mathbf{Q_m} = \mathbf{H}^{(l-1)}\mathbf{W}_{Q,m}, 
\quad \mathbf{K_m} = \mathbf{H}^{(l-1)}\mathbf{W}_{K,m}, 
\quad \mathbf{V_m} = \mathbf{H}^{(l-1)}\mathbf{W}_{V,m} 
\end{equation}

The projected matrices in Eq.~\ref{eq:sa_qkv} are of $N \times d_k$ dimension, where $\mathbf{W}_{Q,m} ,\mathbf{W}_{K,m}, \mathbf{W}_{V,m} \in \mathbb{R}^{d_{\text{model}} \times d_k}$ are learnable projection weights matrices for head $m$ and $d_k = \frac{d_{\text{model}}}{M}$ denotes a per head dimension. 
The query ($\mathbf{Q}_m$) and key ($\mathbf{K}_m$) matrices are learned projections that determine the effect of each singularity on the others, while the value $\mathbf{V}_m$ controls the information exchanged between them. Thus, instead of hard-coding the dynamics of physical singularities, the model can implicitly discover complex physical behaviors among singularities from data. For example, in the case of shocks, the heads can learn to maintain the mutual vertical spacing that determines the viscosity. As such, through self-attention, for each head $m$, the interaction matrix is 
$$\mathbf{A}_m(\mathbf{H}^{(l-1)}) = \operatorname{softmax}\left(\frac{\mathbf{Q}_m\mathbf{K}_m^T}{\sqrt{d_k}}\right).$$
The resulting attention matrix $\mathbf{A}_m$, allows the $N$-singularity tokens to coordinate their relative latent geometry internally. 
The corresponding self-attention update is obtained by applying this interaction matrix to the value matrix:
$$\mathbf{SA}_m(\mathbf{H}^{(l-1)}) = \mathbf{A}_m(\mathbf{H}^{(l-1)})\mathbf{V}_m \in \mathbb{R}^{N \times d_k}.$$
This operation passes information between singularity tokens for head $m$. It allows each singularity to update its latent state based on the other singularities before conditioning on the external observation. 

These $M$ head outputs are then concatenated and projected back to the model dimension (detailed in Sec.~\ref{sec:output}) by a learnable matrix $\mathbf{W}_O^{\text{SA}}$, resulting in the combined self-attention representation: 
\begin{equation}
\label{eq:sa_full}
\mathbf{H}_{\text{SA}} = \operatorname{concat}(\mathbf{SA}_{1}, \dots, \mathbf{SA}_{M}) \mathbf{W}_O^{\text{SA}} \in \mathbb{R}^{N \times d_{\text{model}}}.
\end{equation}
This projection ensures the concatenation restores the full model dimension of size $d_\text{model}$.

\subsubsection*{Generative Conditioning via Cross-Attention}
The described self-attention module provides a mechanism for learning a stable physical geometry among the $N$ individual singularities. 
To generate meaningful signals for a given task, the denoising process must be steered by an external degraded or partial observation, represented as grid samples of signal amplitudes. This grid-based observation does not carry its underlying singularity structure. 
Whether dealing with an audio waveform or a sensor measurement, the generated singularities must reflect the configuration of these dense signals. We define the raw external observation as a dual-channel tensor $\mathbf{U} \in \mathbb{R}^{2 \times S}$. The first channel $\mathbf{c} \in \mathbb{R}^S$ is an external real-valued dense observation, and the second channel contains the corresponding coordinate values over the $S$ sampled locations. The first channel is processed by a standard 1D Convolutional Neural Network (CNN), from which we extract the observation features. This operation results in a physical feature matrix $\mathbf{F}_c\in \mathbb{R}^{S \times (d_{\text{model}}/2)}$ that holds various modality-specific features, such as localized shocks and audio transients. 

The CNN encoder is translation-equivariant; thus, processing the observation signal through it does not preserve absolute positional information. 
Instead, we project the coordinate channel through a coordinate encoder to preserve the grid's absolute addressing. 
This provides a distinct geometric feature matrix $\mathbf{F}_g \in \mathbb{R}^{S \times (d_{\text{model}}/2)}$ encoding the spatial location of the signal features. Combining these two decoupled features, $\mathbf{F}_g$ and $\mathbf{F}_c$ are fused into a Layer Normalized~\cite{ba2016layernormalization} concatenated tensor along the feature dimension, so each localized feature pairs with its spatial coordinate as follows:  
\begin{equation}
\label{eq:ca_fusion}
\mathbf{X} \triangleq \operatorname{LN}\left( [\mathbf{F}_c \,\|\, \mathbf{F}_g] \mathbf{W}_{\operatorname{fuse}} \right),
\end{equation}
where $\operatorname{LN}(\cdot)$ denotes LayerNorm and $\mathbf{W}_{\operatorname{fuse}} \in \mathbb{R}^{d_{\text{model}} \times d_{\text{model}}}$ is a learnable fusion projection.
The resulting fused representation in Eq.~\ref{eq:ca_fusion} serves as an external interface for the cross-attention module. 
Despite our CNN-based encoder being designed to extract steering features from the conditional observation, the network's off-grid property is preserved. 
The rest of the network remains off-grid, with the final generated output represented in the continuous singularity space.

We use Eq.~\ref{eq:sa_full}, the combined multi-head intermediate matrix returned by the previous self-attention module. 
Since $\mathbf{H}_{SA}$ is the internally coordinated singularity state, we define the query as $\mathbf{Q}^{CA}_m = \mathbf{H}_{SA}\mathbf{W}^{CA}_{Q,m}$. The fused external observation in Eq.~\ref{eq:ca_fusion} is used to define the keys and values: $\mathbf{K}^{CA}_m = \mathbf{X}\mathbf{W}^{CA}_{K,m}$ and $\mathbf{V}^{CA}_m = \mathbf{X}\mathbf{W}^{\text{CA}}_{V,m}$. 
The query $\mathbf{Q}^{CA}_m$ holds the internal arrangement computed by the previous self-attention layer. This allows each singularity token to independently retrieve the encoded features $\mathbf{F}_c$ in $\mathbf{X}$. The cross-attention for head $m$ is computed through dot-product attention:
\begin{equation}
\label{eq:ca_softmax}
\mathbf{H}^{\text{CA}}_m =\mathrm{softmax}\left(\frac{\mathbf{Q}^{\text{CA}}_m(\mathbf{K}^{\text{CA}}_m)^\top}{\sqrt{d_k}}\right)
\mathbf{V}^{\text{CA}}_m,
\end{equation}
this results in a learned alignment matrix between the $N$ coordinated singularities and the external observation features $\mathbf{X}$ in Eq.~\ref{eq:ca_fusion}. The softmax normalization in Eq.~\ref{eq:ca_softmax} converts this into an attention distribution, dictating how strongly each of the $N$ singularities relates to the localized features at each of the $S$ coordinates.

\subsubsection*{Feed-Forward Network and Output Projection}
\label{sec:output}
We integrate the individual heads into a unified latent representation by concatenating them together and projecting them back into the latent dimension $d_{\text{model}}$ through a linear transformation: 
$$\mathbf{H}^{\text{CA}} = \operatorname{concat}(\mathbf{H}^{\text{CA}}_1, \dots, \mathbf{H}^{\text{CA}}_M)\mathbf{W}_{O}^{\text{CA}},$$
where $\mathbf{W}_O^{CA} \in \mathbb{R}^{d_{\text{model}} \times d_{\text{model}}}$ is a learnable weight matrix used to combine the information of each independent head. This fusion ensures that features extracted from the external signal are integrated back into a unified representation for each singularity; these are essentially a series of linear operations. To provide a non-linear transformation and learn the residual difference, we employ an FFN that consists of two linear layers with an inner non-linear activation (SiLU): 
$$\mathbf{H}_{\text{out}} = \mathbf{H}^{\text{CA}} + \operatorname{FFN}(\operatorname{LN}(\mathbf{H}^{CA})),$$
where a pre-norm operation is applied to ensure variance stability during training.
Finally, the updated singularity matrix $\mathbf{H}_{\text{out}}$ is processed by a linear decoding head that maps each of the $N$ tokens from the $d_{\text{model}}$ latent dimension back into the coordinate $[\operatorname{Re}(z),\log(\operatorname{Im}(z)),\operatorname{Re}(r),\operatorname{Im}(r)]^\top$, representing the network prediction of the singularity configuration, given the noisy state at timestep $t$.

\subsection{Training Objective and Direct Sample Prediction}
The generative task is defined as reversing the forward SDE defined in Eq.~\ref{eq:diffusion_forward}, which requires learning a score function $s_\theta(\mathcal{Q}_t, t, \mathbf{c})$. Instead of parameterizing the process as noise ($\epsilon$) or velocity ($v$) prediction, as is typically done, we formulate our network and parametrize $s_\theta$ via direct sample prediction - $\hat{Q}_0$-prediction. 
Recent research demonstrates that in various applications, predicting clean data (known $x_0$-prediction) may provide better results than $\epsilon$ or $v$ prediction, as clean data inherently resides on a low-dimensional manifold~\cite{li2026basicsletdenoisinggenerative}. Our choice of $\hat{\mathbf{Q}}_0$-prediction is primarily driven by the geometric and physical constraints of the singularity domain. The Singularity Space configuration is an unordered, permutation-invariant set. This property allows us to compute the optimal bipartite matching between our ground-truth singularity sets and the network's predictions. 
The bipartite matching operates in singularity space, as matching cannot be performed on noise ($\epsilon$), so direct $\hat{\mathbf{Q}}_0$-prediction is needed. 
To enforce our well-defined valid domain $\Omega_{\delta}$, our projection operator $\mathcal{P}_{\Omega_\delta}$ must actively compute pairwise singularity-space distances ($|\zeta(q_i)-\zeta(q_j)|$) at every sampling step. Therefore, predicting absolute coordinates allows direct computation of the distance-minimization projection. Finally, by predicting singularity coordinates at each training step, we retain the flexibility needed to introduce generic auxiliary terms in our loss, such as a physics-informed loss. Thus, practically, we denote $f_\theta(\mathcal{Q}_t, t, \mathbf{c})$ as our network (represents $s_\theta$), which outputs the direct sample prediction $\hat{\mathbf{Q}}_0$. 
We optimize the network through an $L_1$ bipartite matching defined as:
\begin{equation}
\label{eq:l1_training_loss_singularity}
\mathcal{L_\mathcal{Q}} = \mathbb{E}_{t, \mathbf{Q}_0, \boldsymbol{\epsilon}} \left[ \min_{\pi \in \Pi} \frac{1}{N} \sum_{i=1}^N \left\| \hat{\mathbf{q}}_{0,i} - \mathbf{q}_{0,\pi(i)} \right\|_1 \right],
\end{equation}
where $\hat{\mathbf{q}}_{0,i}$ is the $i$-th prediction of $f_\theta$, and $\pi \in \Pi$ assigns each network prediction to a ground-truth singularity $\mathbf{q}_{0,\pi(i)}$. For this assignment problem, denoted as $\pi$, we utilize the Hungarian matching~\cite{hungarian55} algorithm. 

Since our network predicts the clean coordinates $\hat{\mathbf{Q}}_0$ directly and our decoder $\mathcal{D}$ is differentiable in those coordinates, we add a generic (weak) reconstruction term in signal space, so gradients flow back through $\mathcal{D}$ to refine the coordinates. We reconstruct the predicted singularity set and compare it to the ground-truth signal:
\begin{equation}
\label{eq:l2_training_loss_signal}
\mathcal{L}_{\text{signal}} \triangleq \big\| \mathcal{D}(\hat{\mathbf{Q}}_0) - \mathbf{u} \big\|_2^2,
\end{equation}
where $\mathcal{D}$ is the meromorphic decoder that maps a singularity-coordinate configuration to its reconstructed signal by evaluating the
additive expansion in Eq.~\ref{eq:add_rep_mittag_part}. The term (Eq.~\ref{eq:l2_training_loss_signal}) lets us supervise in signal space while generating in singularity space. It should be noted that this term does not hold any problem-specific structure (unlike a PINN residual) and only refines a generic signal reconstruction, so it can be transferred to other modalities and problems. 

Finally, the total dual space objective is defined by combining the two equations, Eq.~\ref{eq:l1_training_loss_singularity} and Eq.~\ref{eq:l2_training_loss_signal}, resulting in the composite loss:
$$\mathcal{L} \triangleq \mathcal{L}_{\mathcal{Q}} + \lambda(e)\,\mathcal{L}_{\text{signal}}, \qquad \lambda(e) \triangleq  \lambda_{\max}\frac{e}{E}, \quad \lambda_{\max} \triangleq 10^{-2}$$ where $e$ is the current epoch and $E$ is the total number of epochs, with $e \leq E$.
Because the Singularity Space coordinates are too noisy early in training for decoder supervision to help, the $\lambda(e)$ term delays the reconstruction contribution until the predicted coordinates are accurate enough for the decoded signal to provide a useful gradient.  

\subsection*{Sampling and Reconstruction: Constrained Multi-Step Inference}
The singularity generation process is performed by reversing the forward SDE defined in Eq.~\ref{eq:diffusion_forward}. This process starts with a Gaussian prior $\mathcal{Q}_T \sim \mathcal{N}(0, \mathbf{I})$ and iteratively reverses the process, converging to the physical clean coordinate $\mathcal{Q}_0$.
As introduced in Sec.~\ref{sec:diffusion}, we implement the continuous reverse process by discretizing the unconstrained SDE, practically utilizing the standard DDPM sampling framework~\cite{ho2020denoisingddpm}:
$$\tilde{\mathcal{Q}}_{t-1} = \frac{\sqrt{\bar{\alpha}_{t-1}}(1 - \alpha_t)}{1 - \bar{\alpha}_t} \hat{\mathcal{Q}}_0 + \frac{\sqrt{\alpha_t}(1 - \bar{\alpha}_{t-1})}{1 - \bar{\alpha}_t} \mathcal{Q}_t + \sigma_t \mathbf{z},$$
where $\alpha_s = 1-\beta_s$  is the per-step retention factor and $$\bar{\alpha}_t = \prod_{s=1}^t \alpha_{s},$$ is the cumulative product over steps, $\beta_s$ is the per-step variance schedule and $\sigma_t$ the reverse-step noise scale. The noise $\mathbf{z} \sim \mathcal{N}(\mathbf{0}, \mathbf{I})$ is injected at each discrete step $t$. The coefficients in this equation act as a weighting mechanism between the network's prediction $\hat{\mathbf{Q}}_{0}$, and the current state $\mathcal{Q}_{t}$. Due to injected stochastic noise, the singularity configuration at each iteration is not guaranteed to remain physically valid. As described in Sec.~\ref{sec:diffusion}, we employ a projection-based method to constrain our reverse process to stay within the physical domain $\Omega_{\delta}$: 
$$\mathcal{Q}_{t-1} = \mathcal{P}_{\Omega_\delta}(\tilde{\mathcal{Q}}_{t-1}).$$
This ensures that the final singularity configuration remains within the valid physical domain $\Omega_\delta$ defined in Eq.~\ref{eq:enforce_op} and is consistent with the learned distribution.

\section{Case Study: Viscous Burgers Equation}
\subsection{Setup}
We empirically evaluate our singularity space representation and generative capabilities on a synthetic dataset of randomly parametrized stationary Burgers shock equations, including their low-viscosity limits. This equation is a fundamental model for studying shock waves across domains, including fluid dynamics, traffic flow, and acoustics. 
The Burgers equation is characterized by its challenging near-singular discontinuities. It is formally defined by:
\begin{equation}
\label{eq:burgers_pde}
\frac{\partial u}{\partial t} + u \frac{\partial u}{\partial x} = \nu \frac{\partial^2 u}{\partial x^{2}}.
\end{equation}
Burgers equation models the competition between two opposing physical forces.
The non-linear advection, $uu_x$, attempts to form a discontinuity, while the viscous diffusion term, $\nu u_{xx}$, attempts to smooth it. In low viscosity settings, where $\nu \to 0$, the solution approaches a discontinuous shock wave, posing a challenge for traditional numerical and neural solvers. While our framework can be applied to various domains with highly transient signals, we first examine its representation of sharp shocks, where we can test its representation capacity as well as its limitations.

\subsection{Synthetic Data Generation}
Real-world datasets typically do not provide explicit singularity annotations for signals. The extraction of singularities from raw signals often introduces numerical errors unrelated to the capacity of our Singularity Space framework. To isolate the evaluation of the representation capacity from these numerical errors, we generate our dataset of singularities using a reverse synthetic data-generation process. Instead of extracting singularities from a signal, specifically shocks, we construct each signal from a known singularity set. This results in a dataset of numerically exact, noise-free singularities used for training supervision, while the model is conditioned only on partial or degraded grid samples of that signal.

\paragraph{Stationary Shock Solution.}
We begin our formulation with the one-dimensional viscous Burgers equation defined as
$$\frac{\partial u}{\partial t} + u \frac{\partial u}{\partial x} = \nu \frac{\partial^2 u}{\partial x^2}$$ 
subject to the boundary conditions:
$$\lim_{x \to -\infty} u(x) = +u_0 \quad \text{and} \quad \lim_{x \to +\infty} u(x) = -u_0$$ 
where $\nu > 0$ is the viscosity and $u_0 > 0$ determines the shock strength.
Our objective is to evaluate representational capabilities in the presence of sharp topological structures, rather than their ability to generate trajectories of "smooth" temporal evolution. Hence, we first construct an exact ground-truth solution for a stationary shock. We assume $\partial u / \partial t = 0$, so the system is reduced to the following non-linear Ordinary Differential Equation (ODE):
$$u \frac{du}{dx} = \nu \frac{d^2u}{dx^2}.$$
Integrating with respect to $x$ we get $\frac{1}{2}u^2 = \nu \frac{du}{dx} + C_{1}$. Applying our boundary conditions (i.e., $u \to \pm u_0$ where spatial gradients, $\frac{du}{dx} \to 0$), we resolve the integration constant $C_1 = \frac{1}{2}u_0^2$, and plugging it back into the equation, we get the separable ODE: 
$$\frac{du}{u^2 - u_0^2} = \frac{dx}{2\nu}.$$
Applying partial fraction decomposition and integrating, we finally get the classic stationary shock profile~\cite{g_b_whitham_linear_1974}:
$$u(x) = -u_0 \tanh\left(\frac{u_0(x-x_c)}{2\nu}\right),$$
where the integration constant determines the shock center $x_c$.

\paragraph{Singularity-Based Representation.}
To generate data through the underlying singularity structure in the complex plane, we use the stationary shock solution of the Burgers equation. The hyperbolic tangent function defined as, $$\tanh(x) = \frac{\sinh(x)}{\cosh(x)},$$ is meromorphic with simple poles of the zeros of $\cosh(x)$. 
According to Mittag-Leffler's theorem, a meromorphic function is uniquely determined by its principal parts at its poles (Sec.~\ref{sec:additive_rep}). 
This allows us to reconstruct the shock profile entirely from its singular points using Mittag-Leffler's tangent expansion:
$$\pi\tan(\pi z) = \lim_{N\to\infty}\sum_{n=-N}^N \frac{-1}{z-(n+\frac{1}{2})}.$$ 
To transform to the hyperbolic tangent form, we first apply the identity $\tanh(x) = -i\tan(ix)$ and substitute $z = \frac{ix}{\pi}$ into the expansion; we get
$$\tanh(x) =  \frac{-i}{\pi} \sum_{n=-\infty}^{\infty} \frac{-1}{\frac{ix}{\pi} - (n+\frac{1}{2})} = \sum_{n=-\infty}^{\infty}\frac{1}{x-i\pi(n+\frac{1}{2})}.$$ 
We apply this expansion to the previously derived shock profile, $u(x) = -u_0 \tanh(\frac{u_0(x-x_c)}{2\nu})$. 
Substituting the expansion and rearranging: 
$$u(x) = -u_0 \sum_{n=-\infty}^{\infty}\frac{1}{\frac{u_0 (x-x_c)}{2\nu}-i\pi(n+\frac{1}{2})} = \sum_{n=-\infty}^{\infty}\frac{-2\nu}{x-z_n},$$
where $z_n = x_c + i\pi \frac{2\nu}{u_0}(n + \frac{1}{2})$, with the residue for every pole set as viscosity. 
This solution provides the final representation that enables the generation of numerically stable shocks.
We keep the $K=32$ upper-half singularities closest to the real axis, as these contribute most to the signal's high-frequency content, and include their conjugates during reconstruction. Thus, the learned representation contains $K=32$ singularity tokens, while the reconstructed meromorphic function contains $2K=64$ poles in total.

\paragraph{Sampling.} 
We randomly sample the three parameters, $\nu \sim \operatorname{LogUniform}(0.002,0.01)$, shock strength, $u_0 \sim \mathcal{U}(0.4, 0.9)$, and the shock center of the spatial domain $x_c \sim \mathcal{U}(-0.8, 0.8)$. 
For each parameter, we compute the corresponding pole configuration and reconstruct the signal on a uniform grid $G=1024$ over $[-1, 1]$. 
Although our synthetic Burgers shocks are generated from three physical parameters, the model is never given the parameters $(\nu,u_0,x_c)$ directly. Instead, it must discover the low-dimensional manifold of valid singularity configurations within this high-dimensional space. 
Successful reconstruction requires generating 32 singularities, corresponding to a \(4K=128\)-dimensional real output. Our synthetic training set contains 4000 shocks, and the test and evaluation sets contain unseen 200 shocks, with a random seed for reproducibility.

\section{Experimental Results}
The primary objective of our experiments is to validate our Singularity Space framework for systems characterized by transients and sharp discontinuities. We perform various degradations, such as down-sampling a high-fidelity signal to $M$ spatial points, creating a degraded observation with partial information near the shock transition. In this setting, the model's task is to recover a physically consistent approximation of the original high-fidelity signal, conditioned on the degraded or partial observation. We tested on both in-distribution and out-of-distribution conditioning, where out-of-distribution (OOD) refers to sampling densities $M$ or Gaussian noise settings, unseen during training and applied only at test time.

We compare against FNO~\cite{li2021fourierneuraloperatorparametric} as a representative Fourier-based dense-output neural baseline that maps the degraded observation directly to a grid-valued reconstruction. FNO is a widely known Fourier-based dense-output baseline, recognized for its resolution-invariant capabilities. It was introduced as a zero-shot super-resolution model that operates on a fixed number of low-frequency Fourier modes. This makes it a relevant benchmark for evaluating our high-fidelity signal recovery capabilities. 
Recent work~\cite{sakarvadia2026falsepromisezeroshotsuperresolution} has shown that the assumed zero-shot multi-resolution capability of FNO can be sensitive to changes in inference resolution, leading to accuracy degradation. In contrast, Singularity Space predicts a sparse singularity configuration that is decoded analytically. The comparison isolates the output representation, specifically, the explicit singularity coordinates versus a dense grid-valued reconstruction in transient-based settings, which is common across domains. Moreover, the comparison tests whether an explicit singularity-based output provides an advantage over grid reconstruction in sharp-transient regimes. Thus, we do not frame our experiment as a general neural-operator benchmark, but as a comparison of output representations under transient-dominated inverse reconstruction. 

Additionally, we compare with the Adaptive Antoulas-Anderson (AAA)~\cite{Nakatsukasa_2018}. AAA is a classical rational approximation algorithm that directly fits a pole-residue representation from observed samples. AAA is an analytical method and produces singularities without any learned prior. We provide AAA with the same conditioning observation as for the other models and limit it to the same singularity capacity ($32$) to match our singularity budget. 

\paragraph{Experimental Setup.} 
All datasets are partitioned into independent training, validation, and testing sets. To ensure reproducibility of our experimental runs, including the on-the-fly data generation used during pre-training, we use a fixed random seed. The generated data is subsequently augmented according to the specific requirements of each experiment. To ensure a fair baseline comparison, hyperparameters were independently tuned for each architecture. The experiments are implemented in Python (v3.10.19) using PyTorch (v2.3.0) with CUDA (v12.1) and trained on a single NVIDIA RTX 4090/3070 GPU. The Singularity Space transformer is optimized using AdamW with an initial learning rate of $2 \times 10^{-3}$, gradient clipping at norm $1.0$, and a cosine annealing schedule. The AAA baseline uses the official SciPy implementation (v1.15.3). The FNO model implementation relies on the official Neural Operator library (v2.0.0)~\cite{neuraloperatorFNO} and is optimized through AdamW with an initial learning rate of $1 \times 10^{-3}$ decayed via a StepLR scheduler $\gamma=0.5$. Both models (FNO, Singularity Space) are trained with a batch size of $32$ and an early stopping patience of 50 epochs. We monitor the relative $L_2$ error of the reconstructed physical signal at the end of each epoch, keeping the checkpoint with the lowest validation error for final inference. 
 
\subsection{In-Distribution Inverse Reconstruction}
\label{sec:indist_inverse_recon}
This experiment's objective is to test our model's ability to reconstruct in-distribution degraded observation signals. The degraded observations denoted $u_{\text{low}}$, are generated by downsampling the original signal $u(x)$ at $M$ random spatial locations. For compatibility with the grid-based FNO input, which requires a full grid, we linearly interpolate $u_{\text{low}}$ back to the grid ($G=1024$). Both models are trained and evaluated on the densities $M \in \{16, 32, 64\}$. 

\begin{table}[H]
\centering
\caption{In-distribution reconstruction for $M \in \{16,32,64\}$. 
Lower is better for $L_1$, $L_2$, and $L_\infty$; TV ${\to 1}$ is better.}
\label{tab:indist}
\begin{tabular}{llrrrrr}
\toprule
Model & \(M\) & \(L_1\) & \(L_2\) & \(L_\infty\) & TV ratio & X-off \\
\midrule
FNO  & 16 & 0.00051 & 0.00130 & 0.0106 & 1.013 &  -- \\
AAA  & 16 & 0.0525  & 0.225   & 1.519  & 1.567 &  -- \\
Singularity & 16 & 0.00079 & 0.00306 & 0.0243 & 1.000 &  0.00033 \\
\midrule
FNO  & 32 & 0.00033 & 0.00075 & 0.0065 & 1.005 &  -- \\
AAA  & 32 & 0.0233  & 0.191   & 3.049  & 2.999 &  -- \\
Singularity & 32 & 0.00065 & 0.00235 & 0.0189 & 1.000 &  0.00024 \\
\midrule
FNO  & 64 & 0.00026 & 0.00056 & 0.0048 & 1.003 &  --    \\
AAA  & 64 & 0.0080  & 0.0496  & 0.436  & 1.152 &  -- \\
Singularity & 64 & 0.00058 & 0.00204 & 0.0160 & 1.000 &  0.00021 \\
\bottomrule
\end{tabular}
\end{table}

As shown in Table~\ref{tab:indist}, at $M=16$, FNO obtains $L_2=0.00130$, while our model obtains $0.00306$. Both these errors are small. This indicates that the inverse reconstruction problem is well solved in-distribution by both models. TV is preserved across all densities by our model, whereas FNO shows a small but increasing deviation in TV as observation density decreases (from 1.003 at $M=64$ to 1.013 at $M=16$). FNO achieves lower $L_2$ and $L_\infty$ reconstruction errors. Unlike FNO, the singularity space predicts the position error offset, up to $10^{-4}$, which corresponds to about $\frac{1}{7}$ of a grid cell (with $G=1024$). This offset is only a fraction of a grid cell; sharp analytic reconstruction makes pointwise metrics highly sensitive to it. This explains its higher pointwise errors, $L_2$ and $L_\infty$, and also its sensitivity to shock-position error, denoted in Table~\ref{tab:indist} as $\text{X-off} = \mathbb{E}|\hat{x}_c - x_c|$. 

The AAA baseline performs substantially worse than both learned models. As shown in Table~\ref{tab:indist}, when fitted to $u_{\mathrm{low}}$ and evaluated against the target $u(x)$, its error and TV ratio increase substantially. This is expected as it lacks a prior and fits the interpolated degraded observation directly rather than recovering the clean signal. For this specific experiment, the difficulty lies not in the approximation itself, but in inverse recovery from degraded observations. Moreover, to separate the inverse problem and learning capabilities, we found that AAA accurately fits the degraded observation when evaluated against $u_{\text{low}}$. 

\paragraph{Position-aligned diagnostic.} We investigate whether Singularity Space $L_2$ and $L_\infty$ errors are due to structure or location. As shown in Table~\ref{tab:m64_clean_shift_diagnostic}, for $M=64$, we shift the real-coordinate offset by the ground-truth offset pre-reconstruction. This reduces Singularity Space $L_2$ from $0.00204$ to $0.00044$. The position-aligned diagnostic reduces $L_2$ by $4.6\times$, bringing it below the FNO baseline while preserving the signal structure (TV). 

\begin{table}[H]
\centering
\caption{Diagnostic for Singularity Space at $M=64$, evaluated on 200 test samples. For each Singularity Space prediction, we shift the predicted singularity coordinates by applying the ground-truth position offset $\Delta x = x_{\text{gt}} - x_{\text{pred}}$ before reconstruction, thereby isolating shape error from localization error. }
\label{tab:m64_clean_shift_diagnostic}
\begin{tabular}{llrrrrr}
\toprule
Model & $L_1$ & $L_2$ & $L_\infty$ & TV ratio & X-off \\
\midrule
FNO & 0.00026 & 0.00056 & 0.00476 & 1.003 & -- \\
Singularity & 0.00058 & 0.00204 & 0.0160 & 1.000 &  0.00021 \\
Singularity (aligned) & 0.00033 & \textbf{0.00044} & \textbf{0.00198} & 1.00008 & 0.00001 \\
\bottomrule
\end{tabular}
\end{table}
The diagnostic also shows a substantial reduction in $L_1$ and $L_\infty$, while leaving the TV ratio unchanged. This shows that the dominant error is a localization issue rather than a structural distortion of the reconstructed shock. If the generated singularity structure were incorrect, correcting only a global real-coordinate offset would not reduce the pointwise errors to this extent. The diagnostic is not used as a primary metric because it relies on ground truth offsets. We further analyze this failure in the noise robustness experiment.

\subsection{Zero-Shot Multi-Resolution Reconstruction}
In this experiment, we evaluate both models on OOD range - sub-resolution and super-resolution. 
Fourier-based networks exhibit resolution-transfer behavior because they operate on a fixed number of low-frequency Fourier modes. 
A recent study has shown that changing the inference resolution can degrade zero-shot multi-resolution performance~\cite{sakarvadia2026falsepromisezeroshotsuperresolution}. 
In our experimental setting, both models are trained on a multi-resolution set $M \in \{16, 32, 64\}$. The grid remains fixed at $G=1024$, but the conditioning density $M$ is varied outside the training range. We use $M=8$ and $M=128$ to evaluate the robustness to unseen observation densities. 

\begin{figure}[H]
\centering
\includegraphics[width=\linewidth]{figures/singulairty_anim.gif_0.pdf}
\includegraphics[width=0.75\linewidth]{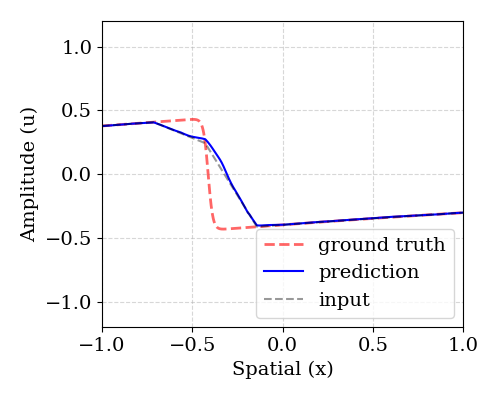}
\caption{Representative zero-shot reconstruction at $M=8$ conditioned on partial observation (grey dashed), outside the training range $M\in\{16,32,64\}$. 
Top left: the predicted coordinates aligned with the ground-truth singularity configuration. Top right: the signal reconstruction (blue) matches the ground truth (red). Bottom: FNO signal reconstruction (blue) on the same degraded input (grey dashed), shown with the ground truth (red).}
\label{fig:m8_zeroshot}
\end{figure}

As shown in Table~\ref{tab:zeroshot}, FNO performs well at the training densities $\{16,32,64\}$ but degrades considerably in accuracy at the unseen densities $M=8$ and $M=128$. This is consistent with previous work~\cite{sakarvadia2026falsepromisezeroshotsuperresolution}, which showed that multi-resolution training mitigates the zero-shot failure within the trained resolution range, but that learned operators typically do not generalize beyond it. 

\begin{table}[H]
\centering
\caption{Zero-shot multi-resolution reconstruction. Test densities 
$M=8$ (sub-resolution) and $M=128$ (super-resolution) lie outside 
the training range $\{16,32,64\}$. Lower is better for $L_1$, $L_2$, $L_\infty$. TV ratio closer to $1.0$ is better}
\label{tab:zeroshot}
\vspace{0.5em}
\begin{tabular}{llrrrrr}
\toprule
Model & $M$ & $L_1$ & $L_2$ & $L_\infty$ & TV ratio & X-off \\
\midrule
FNO  & 8  & 0.0980 & 0.3604 & 1.583 & 1.170 & - \\
Singularity & 8  & \textbf{0.0384} & \textbf{0.0850} & \textbf{0.369} & \textbf{1.003} & 0.0175 \\
\midrule
FNO  & 128 & 0.00082 & 0.00403 & 0.0378 & 1.005 & - \\
Singularity & 128 & \textbf{0.00060} & \textbf{0.00240} & \textbf{0.0194} & \textbf{1.000} & 0.00023 \\
\bottomrule
\end{tabular}
\end{table}

As shown in Table~\ref{tab:zeroshot}, at density $M=8$, FNO degrades with $L_2 = 0.360$, $L_\infty = 1.583$ and the TV ratio reaches $1.170$. 
This is approximately $17\%$ excess over the ideal TV ratio of $1.0$. In contrast, our model demonstrates structural stability under the same zero-shot conditions ($L_1 = 0.0384, L_2 = 0.0850$ and TV ratio is $1.003$), achieving approximately $4.2\times$ lower error while preserving the shock structure with only $0.3\%$ excess variation. At $M=128$, FNO's $L_2$ degrades approximately by $7\times$ relative to its best in-distribution minimum results in Table~\ref{tab:indist}. Our model remains close to its in-distribution error with a robust TV ratio. This suggests that the singularity representation is less sensitive to OOD shifts in observation resolution than the FNO baseline.

\subsection{Decoder Resolution Independence}
The previous experiments, as shown in Table~\ref{tab:indist}, evaluated reconstructions on a fixed grid $G=1024$, while changing the conditioning observation resolution $M$. 
In this experiment, we isolate the decoder property and demonstrate that once a singularity set has been generated, our decoder can reconstruct the analytic representation on different output grids. We test this by fixing the degraded conditioning at $M=64$ from the previous Table~\ref{tab:indist} and reconstructing the same predicted singularity configurations at three unseen output grids $G \in \{512, 2048, 4096\}$, where these reconstruction grids are not part of the learned output representation.

\begin{table}[H]
\centering
\caption{Resolution-free reconstruction of 200 test samples, with $M=64$ on different output grids, without retraining the model. The conditioning observation and predicted singularity configuration were not changed. Only the decoder grid is changed.}
\label{tab:multigrid_rendering}
\begin{tabular}{lrrrr}
\toprule
Output grid $G$ & $L_1$ & $L_2$ & $L_\infty$ & TV ratio \\
\midrule
512  & 0.00060 & 0.0021 & 0.0172 & 1.000 \\
1024 (original) & 0.00058 & 0.00204 & 0.0160 & 1.000 \\
2048 & 0.00061 & 0.0021 & 0.0174 & 1.000 \\
4096 & 0.00061 & 0.0021 & 0.0174 & 1.000 \\
\bottomrule
\end{tabular}
\end{table}

The results in Table~\ref{tab:multigrid_rendering} are stable across various unseen output grids, demonstrating that our reconstructed signal is grid-independent. More specifically, we show that the inferred singularity set defines an analytic meromorphic representation that can be evaluated at various output resolutions without any degradation in the reconstruction.

\subsection{Robustness to Test-Time Noise}
In Table~\ref{tab:indist}, we have demonstrated how both models are robust to lower-resolution reconstruction. We now evaluate the behavior of the inverse reconstruction when the conditional signal is corrupted by various levels of Gaussian noise. Both models are trained only on clean signals. 
At test time, we add Gaussian noise. This noise is added to the conditional signal at $M=64$. We tested three noise levels: $2\%$, $5\%$, and $10\%$ of per-signal RMS, corresponding to SNR levels of 34, 26, and 20 dB, respectively.

\begin{table}[H]
\centering
\caption{Noise robustness at $M=64$: Gaussian noise added to the input observation, scaled to per-signal RMS. Both methods were trained on clean signals only and evaluated on 200 test samples. Lower is better for $L_1$, $L_2$, $L_\infty$, and X-off. TV ${\to 1}$ is better.}
\label{tab:noise_results}
\begin{tabular}{llrrrrr}
\toprule
Noise & Model & $L_1$ & $L_2$ & $L_\infty$ & TV ratio & X-off \\
\midrule
$0\%$  & FNO  & 0.00026 & 0.00056 & 0.00476     &  1.0031 & -- \\
$0\%$  & Singularity & 0.00058 & 0.00204 & 0.0160      &  1.000   &  0.00021 \\
$0\%$  & AAA  & 0.00795 & 0.04956 & 0.43571     &  1.1515 & -- \\
\midrule
$2\%$  & FNO  & 0.01464 & 0.02482 & 0.14748     &  1.1910  & -- \\
$2\%$  & Singularity & 0.07763 & 0.23546 & 1.21495     &  1.0068  & 0.02206 \\
$2\%$  & AAA  & 0.05513 & 0.51986 & 11.80027    & 13.9143 & -- \\
\midrule
$5\%$  & FNO  & 0.03640 & 0.06015 & 0.34847     &  1.7110  & -- \\
$5\%$  & Singularity & 0.16474 & 0.40657 & 1.66074     &  1.0547  & 0.04200 \\
$5\%$  & AAA  & 0.22115 & 4.12978 & 92.82689    & 79.7851 & -- \\ 
\midrule
$10\%$ & FNO  & 0.07159 & 0.10882 & 0.56011     & 2.6536   & -- \\
$10\%$ & Singularity & 0.28874 & 0.58231 & 1.94370     & 1.1839   & 0.06275 \\
$10\%$ & AAA  & 0.33626 & 5.32724 & 122.50533   & 109.9134 & -- \\
\bottomrule
\end{tabular}
\end{table}

\begin{figure}[H]
  \centering
  \includegraphics[width=0.7\linewidth]{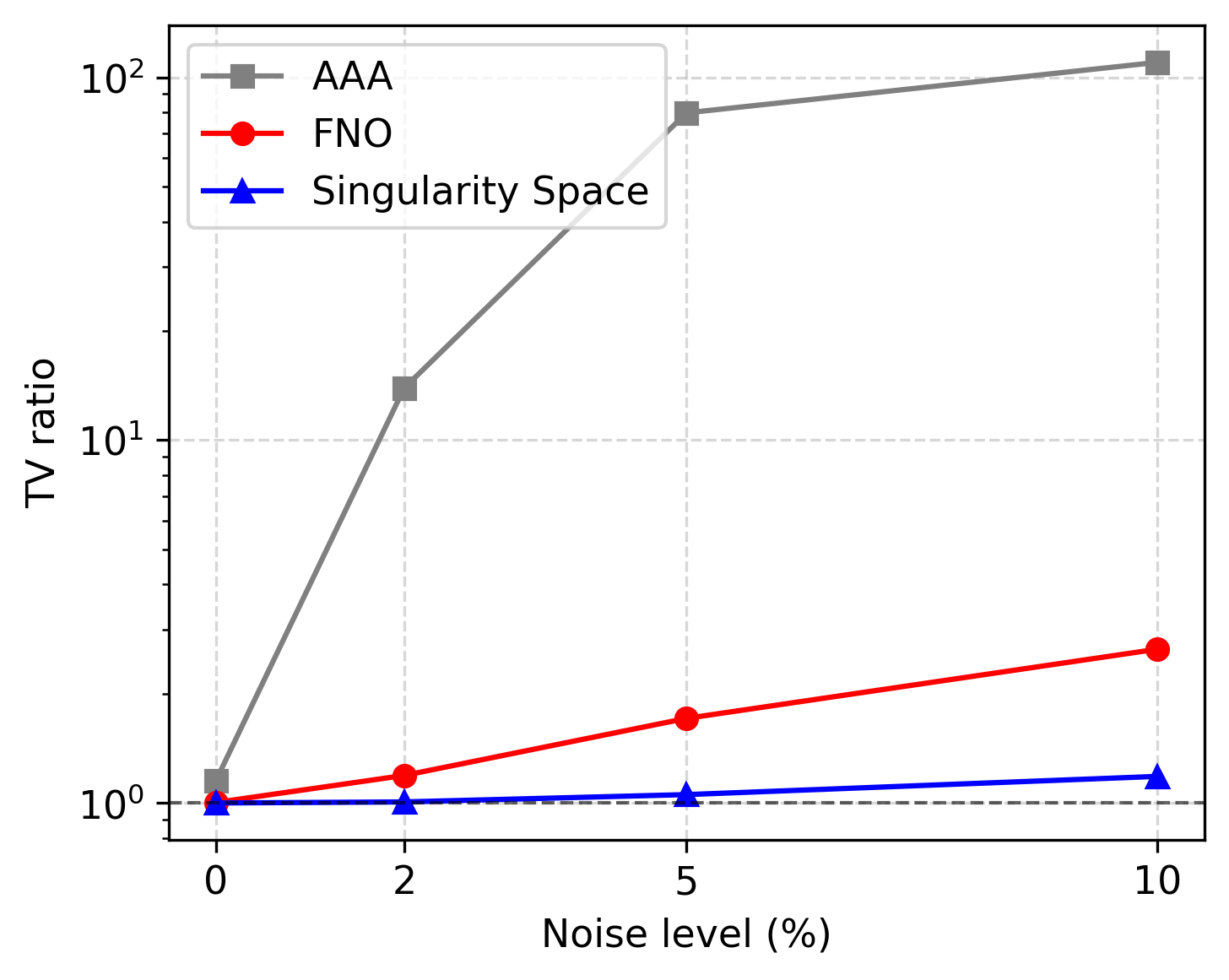}
  \caption{TV ratio versus test-time noise level at $M=64$ (log scale). AAA diverges and FNO's TV ratio grows to $2.65$ at $10\%$ noise, while Singularity Space remains close to the ideal value of $1.0$ (dashed line), indicating that the shock structure is preserved under unseen observation noise.}
  \label{fig:tv_noise}
\end{figure}

As shown in Table~\ref{tab:noise_results}, FNO achieves lower pointwise errors under unseen noisy conditioning, but its TV ratio increases with added noise, indicating non-physical variation near the shock. More specifically, FNO predicts a signal that includes the input's noise. This behavior is expected, as neither model is trained on noise. FNO is a direct observation-to-signal map with a dense grid-valued output, so it can struggle to separate perturbations from signal. The Singularity Space reverse diffusion is constrained to structurally valid pole configurations, so the reconstruction preserves signal structure despite the corruption. Thus, Singularity Space maintains a TV ratio much closer to unity, suggesting that the analytic shock is structurally preserved. Figure~\ref{fig:tv_noise} demonstrates this trend across noise levels: the singularity reconstruction remains near the ideal TV ratio, while both baselines degrade.

As stated in Sec.~\ref{sec:indist_inverse_recon}, AAA has no prior and becomes unstable despite its excellent performance on clean signals. Moreover, in contrast to the setting in Sec.~\ref{sec:indist_inverse_recon}, where AAA fit the degraded observation $u_{\text{low}}$ accurately and failed only on the clean target, here AAA's reconstruction fails on both the clean signal and its degraded observation $u_{\text{low}}$. Specifically, at $M=64$, adding only $2\%$ noise increases the TV ratio from $1.15$ to $13.91$, and $L_\infty$ from $0.44$ to $11.80$. 
The same trend becomes more severe as the noise level increases. This indicates that rational fitting amplifies noise, whereas the learned singularity prior yields physically correct shock profiles.

\paragraph{Position-aligned diagnostic.} We observe a similar behavior to Sec.~\ref{sec:indist_inverse_recon}, where the error is primarily caused by a small sub-grid localization offset (approximately $1/7$ of a grid cell). To validate this interpretation, we once again isolate the position from the shape error (as in Table~\ref{tab:m64_clean_shift_diagnostic}). We align the predicted singularity set in Table~\ref{tab:noise_results} for $M=64$ with noise of 10\%, shifting the predicted singularity by the true position offset $\Delta x = x_{\text{gt}} - x_{\text{pred}}$. The resulting reduction demonstrates that the pointwise failure is a localization error rather than a structural one.

\begin{table}[ht]
\centering
\caption {Diagnostic at $M=64$ with $10\%$ test-time noise, evaluated on 200 test samples. FNO is shown for reference. The raw Singularity Space prediction is compared with a diagnostic variant in which the predicted singularity set is shifted by the ground-truth real-coordinate offset before reconstruction.}
\label{tab:shift_aligned_diagnostic_noise}
\begin{tabular}{lrrrrr}
\toprule
Model & $L_1$ & $L_2$ & $L_\infty$ & TV ratio & X-off \\
\midrule
FNO    & \textbf{0.07159} & \textbf{0.10882} & 0.56011 & 2.65365 & -- \\
Singularity   & 0.28874 & 0.58231 & 1.94370 & \textbf{1.18399} & 0.06275 \\
Singularity (aligned) & 0.12725 & 0.14464 & \textbf{0.28290} & 1.18734 & \textbf{0.00014} \\
\bottomrule
\end{tabular}
\end{table}

As shown in Table~\ref{tab:shift_aligned_diagnostic_noise}, this position alignment yields $L_2$ decrease from $0.58231$ to $0.14464$, and $L_\infty$ decreases from $1.94370$ to $0.28290$. 
This confirms that much of the pointwise error is due to shock misalignment. The TV ratio remains similar after our positional alignment, indicating that the remaining TV deviation is not position-related. This makes Singularity Space competitive with FNO at $L_2$ and better in $L_\infty$, while maintaining physical structure with a lower TV ratio. 
This alignment is for diagnostic purposes only, since it uses the target shock position after inference.

\subsection{Reconstruction Interpretability}
The interpretability of our model is inherent in the singularity representation itself. In our setting, any meromorphic signal can be represented by the locations and residues of its poles. The generated poles are not abstract latent vectors, but hold physical meaning. For the Burgers shock, each pole $z_k$ has a direct physical interpretation. The real part is the shock position $x_c$, the imaginary part encodes the ratio $\nu/u_0$, and the residue encodes the viscosity through $-2\nu$. Generating the correct set of singularities represents the true physical parameters of the shock. 

\begin{table}[H]
\centering
\caption{
A representative singularity recovery for a test shock at $M=32$ and $\nu = 5.71\times 10^{-3}$. Given a degraded shock, our model predicts 32 upper-half-plane poles of the physically consistent shock. First and last four complex coordinates are shown.}
\vspace{0.5em}
\label{tab:shock_singularity}
\begin{tabular}{clccc}
\toprule
Index & Quantity & Target & Predicted & Abs. error \\
\midrule
All & $x_c$ & $-0.4123$ & $-0.4121$ & $2.0\times 10^{-4}$ \\
\midrule
1  & $\mathrm{Im}(z_1)$  & $0.0408$ & $0.0408$ & $<10^{-4}$ \\
2  & $\mathrm{Im}(z_2)$  & $0.1224$ & $0.1224$ & $<10^{-4}$ \\
3  & $\mathrm{Im}(z_3)$  & $0.2039$ & $0.2039$ & $<10^{-4}$ \\
4  & $\mathrm{Im}(z_4)$  & $0.2855$ & $0.2855$ & $<10^{-4}$ \\
\multicolumn{5}{c}{$\vdots$} \\
29 & $\mathrm{Im}(z_{29})$ & $2.3250$ & $2.3250$ & $<10^{-4}$ \\
30 & $\mathrm{Im}(z_{30})$ & $2.4065$ & $2.4066$ & $1.0\times 10^{-4}$ \\
31 & $\mathrm{Im}(z_{31})$ & $2.4881$ & $2.4882$ & $1.0\times 10^{-4}$ \\
32 & $\mathrm{Im}(z_{32})$ & $2.5697$ & $2.5698$ & $1.0\times 10^{-4}$ \\
\midrule
All & Residue coefficient $-2\nu$ & $-0.0114$ & $-0.0114$ & $<10^{-4}$ \\
\bottomrule
\end{tabular}
\end{table}

Table~\ref{tab:shock_singularity} shows pole recovery for a representative low viscosity shock ($\nu = 5.71\times10^{-3}$). The position is within $2\times10^{-4}$ and pole spacing to four decimals across all 32 poles. Although this physical interpretation is specific to the Burgers solution, the geometric interpretation of the coordinates is a property of the representation itself and generalizes to other signal in the meromorphic class. FNO does not output these physical quantities directly. Its output is a signal from which these quantities cannot be directly recovered. 

The Singularity Space representation is not only interpretable, but compact. Each reconstructed shock is represented by $K$ singularity tokens (here, $K=32$), rather than by 1024 grid values in a spatial reconstruction, which is an $8\times$ reduction in output dimensionality. This compactness property is at the output and representation level and is independent of network size. It should be noted that our inference network has more parameters than FNO (267K vs.\ 198K); however, its per-sample output is smaller. Although FNO operates internally on a truncated Fourier basis, those coefficients are mixed through the network's nonlinearities rather than by an analytic decomposition of the output. Hence, FNO's predicted signals do not factor into physical parameters.

\section{Discussion and Future Work}
We introduced Singularity Space, a meromorphic pole-residue representation for signals with sharp transients or shock-like structures.
We validated the framework on viscous Burgers shocks, where the analytical solution provides exact ground-truth singularities, isolating the representational capacity from data-extraction error. Although a Burgers shock can be characterized by a small number of physical variables, including shock location, viscosity, and amplitude, these parameters are not provided to the model. We learn the distribution of a full $128$-dimensional singularity configuration from external observations and reconstruct the signal analytically. This aligns with the objective of learning singularity-based representations that transfer to other transient-based modalities.

We implemented our Singularity Space by building a conditional diffusion transformer that generates singularity sets rather than dense grid values. By comparing with an FNO baseline, we show that Singularity Space achieves competitive in-distribution reconstruction accuracy while preserving the shock-specific structure that FNO loses in the presence of noise. In the OOD sub-resolution tests, for $M=8$ our framework achieves $L_2=0.0850$, while FNO's $L_2=0.360$, a nearly $4.2\times$ reduction, demonstrating resolution-robust generalization without retraining, consistent with its resolution-free construction. We also demonstrated that the singularity sets are not arbitrary latent variables, but enable an interpretable representation. Specifically, in the Burgers shock setting, their real parts encode the shock location, their imaginary part encodes the viscosity-dependent shock width, and their residues encode the viscosity scale. Thus, the model recovers a compact analytic description of the signal, rather than an abstract latent vector or a dense grid reconstruction. 

The noise experiments (only test-time noise) reveal a limitation of our current framework: its predicted position drifts from the ground-truth value. This yields large $L_2$ and $L_\infty$ pointwise reconstruction penalties, whereas the total variation (TV) indicates preservation of the signal's physical structure. The $L_2$ and $L_\infty$ pointwise sensitivity is mainly related to the prediction of the absolute coordinates in the singularity representation. This suggests that the representation remains structurally correct. In our parameterization, the real component of each pole represents its physical location in the spatial domain. Therefore, a small error in the predicted shock center shifts the real part of the entire pole configuration. While this makes our representation interpretable, it makes it sensitive to global translations. We validated this using position-aligned diagnostics and showed that the pointwise error is due to the shock localization. In other words, the representation separates the error into position and shape, which is typically hard to obtain from a grid output. We investigated several potential sources of this drift, including encoder spatial mixing, sampler behavior, loss-channel balance, and coordinate scaling. The inspection suggests that the offset is not merely an implementation artifact but a generalization-sensitive localization error in the learned inverse map. Since the Singularity Space representation is interpretable, the position accuracy can be mitigated through several approaches, such as test-time optimization~\cite{snell2024scalingllmtesttimecompute, ma2025inferencetimescalingdiffusionmodels} or physics-informed refinement at inference time to handle observational noise robustly. So instead of increasing model capacity to improve position precision, the analytic decoder can serve as a physically grounded verifier for singularity configurations, a direction we leave for future work.  

Our current work focuses on representational capacity, limited to stationary 1D reconstruction. 
Extending Singularity Space to temporal dynamics is an interesting future direction, as the analytic reconstruction of the Singularity Space may overcome the typical temporal accumulated errors of standard neural operators by modeling pole trajectories as latent ODEs. 
Because the DiT architecture dynamically learns relational features rather than relying on physical operators, it may be considered as PDE-agnostic. An interesting direction could be to apply our architecture to other integrable systems governed by singularity dynamics, such as the Korteweg-De Vries (KdV) equation and the Nonlinear Schroedinger (NLS) equations.
Moreover, extending Singularity Space to temporal dynamics would enable the framework to be applied to other signal domains, such as speech and biomedical time series. Speech is one of the more interesting modalities for future directions because these signals contain both smooth oscillatory components and sharp transient events; moreover, they have a long history of pole-based modeling via linear prediction~\cite{MarkelJohnD.1976Lpos}.

One challenge is the absence of ground-truth pole labels during training. This motivates future work on self-supervised mechanisms to learn these representations, either through offline rational-approximation~\cite{vectorfitting_1999, Nakatsukasa_2018} methods to generate pseudo-labels or through self-supervised reconstruction with the analytic decoder. 
Furthermore, extending this approach to higher-dimensional domains, where singularities are no longer isolated points, may require localized patch formulations or Radon-domain decompositions~\cite{coifman2026holomorphic}. Such extensions could enable Singularity Space to be applied to other domains including image processing, vision, and higher-dimensional physical fields. 

Another limitation of our current implementation is that the conditioning observations are still provided on a fixed grid and processed by a convolution network. Thus, the current model separates the output representation from the grid, but not the input conditional signal interface. An extension for this is to improve the encoder with a coordinate-value set encoder that includes sparse signal observations $\{(x_i,u_i)\}_{i=1}^M$. 

\section*{Acknowledgment} AA and EB were partially supported by the Israel Science Foundation (ISF 1873/21), Blavatnik
Computer Science Research Fund, and the School of Mathematical Sciences, Tel Aviv University, Israel.

\bibliographystyle{plain}
\bibliography{references}

\end{document}